\documentclass{article}

\usepackage[preprint]{neurips_2026}
\usepackage[utf8]{inputenc}
\usepackage[T1]{fontenc}
\usepackage{hyperref}
\usepackage{url}
\usepackage{booktabs}
\usepackage{amsfonts}
\usepackage{amsmath}
\usepackage{microtype}
\usepackage{graphicx}
\usepackage{subcaption}
\usepackage{tikz}
\usetikzlibrary{calc}
\usepackage[version=4]{mhchem}
\usepackage{siunitx}
\usepackage{multirow}
\usepackage{xcolor}
\usepackage{caption}
\graphicspath{{resources/}}
\providecommand{\botrule}{\bottomrule}

\usepackage{titlesec}
\titlespacing*{\section}{0pt}{1.6ex plus 0.3ex minus 0.2ex}{0.8ex plus 0.1ex}
\titlespacing*{\subsection}{0pt}{1.2ex plus 0.2ex minus 0.1ex}{0.4ex plus 0.1ex}
\titlespacing*{\subsubsection}{0pt}{1ex plus 0.2ex minus 0.1ex}{0.3ex plus 0.1ex}

\newcommand{\stemgym}{\textsc{STEMGym}}
\newcommand{\decauc}{\text{DEC-AUC}}

\title{\stemgym{}: Benchmarking Sequential Decision-Making under Dose Budgets in Autonomous Electron Microscopy}

\author{%
  \small 
  \begin{tabular}{c@{\hskip 0.6in}c}
    \textbf{\normalsize Can Polat} & \textbf{\normalsize Erchin Serpedin} \\
    \footnotesize Dept. of Electrical \& Computer Eng. & \footnotesize Dept. of Electrical \& Computer Eng. \\
    \footnotesize Texas A\&M Univ. & \footnotesize Texas A\&M Univ. \\
    \footnotesize College Station, TX, USA & \footnotesize College Station, TX, USA \\
    \texttt{\footnotesize can.polat@tamu.edu} & \texttt{\footnotesize eserpedin@tamu.edu} \\
    \\[0.05in] 
    \textbf{\normalsize Mustafa Kurban}\thanks{Corresponding author.} & \textbf{\normalsize Hasan Kurban}\footnotemark[1] \\
    \footnotesize Dept. of Prosthetics \& Orthotics & \footnotesize College of Science \& Eng. \\
    \footnotesize Ankara Univ., Ankara, Turkey & \footnotesize Hamad Bin Khalifa Univ. \\
    \footnotesize \emph{and} Texas A\&M Univ. at Qatar & \footnotesize Doha, Qatar \\
    \footnotesize Doha, Qatar & \texttt{\footnotesize hkurban@hbku.edu.qa} \\
    \texttt{\footnotesize kurbanm@ankara.edu.tr} &
  \end{tabular}
}

\begin{document}

\maketitle

\begin{abstract}
A central premise of autonomous scientific imaging is that smarter \emph{navigation}, whether Bayesian, RL-based, or otherwise adaptive, is the principal lever for sample-efficient acquisition. We present evidence to the contrary in scanning transmission electron microscopy (STEM), an atomic-resolution imaging modality whose every measurement deposits damaging electron dose. We introduce \stemgym{}, an open-source Gymnasium benchmark of 15 physics-simulated STEM worlds spanning five materials, three difficulty levels, and four characterisation tasks, scored by the Dose-Efficiency Curve area (\decauc{}), a single scalar capturing the information-vs-dose Pareto frontier. Across 33 agent configurations under realistic dose budgets, the dominant determinant of dose efficiency is the \emph{analyst} (perception) pipeline, not the navigator: pairing a trained CNN analyst with naïve raster scanning raises \decauc{} by $5.5\times$ over a CNN-free raster baseline ($0.287$ vs.\ $0.052$), while substituting Bayesian or adaptive finite-state-machine navigation for raster yields no statistically significant further gain. Production-tier vision-language models further underperform task-specific CNNs by ${\sim}13\times$ on crystallographic defect analysis. By decoupling perception, navigation, and planning under a unified dose budget, \stemgym{} reframes where ML effort should be invested in autonomous electron microscopy and provides the measurement infrastructure to test it.
\end{abstract}

\section{Introduction}
\label{sec:intro}

Scanning transmission electron microscopy (STEM) is among the most powerful techniques for characterising materials at the atomic scale.
By focusing a sub-\aa ngstr\"om electron probe and raster-scanning across a specimen, STEM produces high-angle annular dark-field (HAADF) images with contrast proportional to $Z^{1.7}$, enabling direct visualisation of atomic columns, point defects, and phase boundaries~\citep{pennycook2011scanning}.
However, the electron probe also damages the specimen: beam-induced knock-on displacement, radiolysis, and heating accumulate with total dose, progressively altering the structure being measured~\citep{egerton2004electron, egerton2021radiation}.
For beam-sensitive materials such as 2D chalcogenides, where the beam itself creates the vacancies under study~\citep{komsa2012two}, dose efficiency is fundamental.

The natural response is autonomous agents that adaptively allocate dose, concentrating measurements where information density is highest. The field has converged on adaptive \emph{navigation} as the principal lever. Yet systematic progress in this direction is hard to track. Agents are reported on incompatible specimens, instruments, dose budgets, and task definitions, with no shared protocol that isolates \emph{which} component of the acquisition pipeline (perception, navigation, planning) is responsible for any observed gain. Cryo-EM has CryoBench~\citep{levy2024cryobench} as a community reference for heterogeneity reconstruction; HAADF-STEM has no analogue, an absence the Microscopy Hackathon~\citep{pratiush2025michackathon} flagged as a primary bottleneck for ML adoption.

\begin{figure}[t]
\centering
\begin{tikzpicture}
\node[inner sep=4pt, fill=gray!6, rounded corners=3pt, draw=gray!25, line width=0.4pt] (overview) {%
\begin{minipage}{0.95\textwidth}
\centering
\begin{minipage}[t]{0.49\textwidth}
\centering
\includegraphics[width=\textwidth]{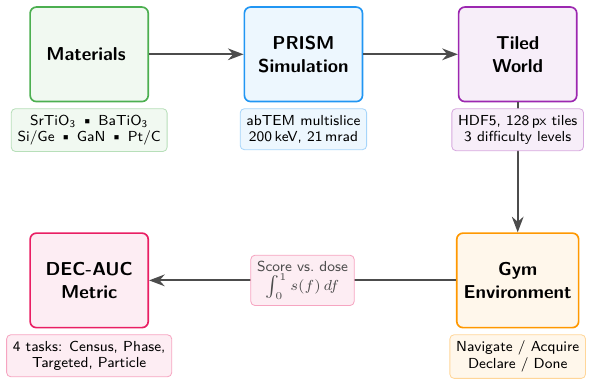}
\end{minipage}\hfill
\begin{minipage}[t]{0.005\textwidth}
\centering
\tikz{\draw[densely dashed, gray!50] (0,0.3cm) -- (0,-3.2cm);}%
\end{minipage}\hfill
\begin{minipage}[t]{0.49\textwidth}
\centering
\includegraphics[width=\textwidth]{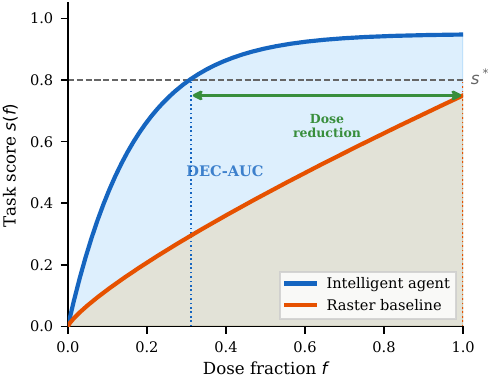}
\end{minipage}
\end{minipage}%
};
\node[font=\bfseries, anchor=south] at ([yshift=-18pt]overview.north west -| {$(overview.west)!0.27!(overview.east)$}) {(a)};
\node[font=\bfseries, anchor=north west] at ([xshift=30pt, yshift=-6pt]{$(overview.north west)!0.52!(overview.north east)$}) {(b)};
\end{tikzpicture}

\vspace{2pt}

\noindent
\begin{tikzpicture}
\node[inner sep=4pt, fill=gray!6, rounded corners=3pt, draw=gray!25, line width=0.4pt] {%
\includegraphics[width=0.95\textwidth]{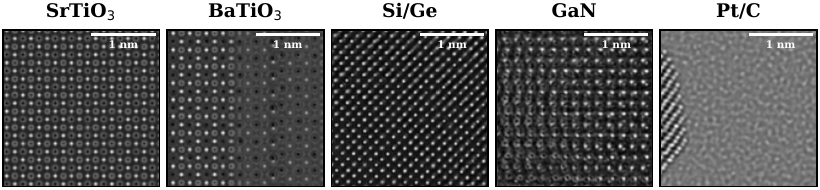}%
};
\end{tikzpicture}
\caption{\stemgym{} overview and material systems.
\textbf{(a)}~Platform pipeline: crystallographic structures are simulated via PRISM multislice into tiled HDF5 worlds; a Gymnasium environment exposes Navigate/Acquire/Declare/Done actions.
\textbf{(b)}~DEC concept: an intelligent agent (blue) reaches accuracy threshold $s^*$ at lower dose fraction than raster (orange); shaded areas represent DEC-AUC.
Bottom: representative HAADF-STEM tiles from each of the five material systems.}
\label{fig:overview}
\end{figure}

We close this gap with \stemgym{}, a Gymnasium~\citep{towers2024gymnasium}-compatible benchmark (Figure~\ref{fig:overview}) that holds dose budget, task definition, and evaluation protocol fixed across agents. This common ground exposes a result that, to our knowledge, has not previously been reported in autonomous STEM and is made visible by a factorial experimental design that has not been applied here before: the dominant determinant of dose efficiency is the \emph{analyst}, not the navigator. Contrary to the community's focus on adaptive acquisition strategies, equipping a trivial raster scan with a trained perception module already closes most of the achievable gap to a fully adaptive agent. Across approximately 8{,}000 episodes, Raster+Analyst achieves Dose-Efficiency Curve area (DEC-AUC) $0.287$ on defect census versus $0.052$ for Raster alone; adding \emph{open-loop} Bayesian navigation or FSM planning on top of the same analyst yields no further statistically significant gain. The same pattern is independently corroborated by the PF GP-BO~\citep{ziatdinov2022hypothesis,kalinin2023ml4stem} and persists across a $50\times$ dose-budget range up to \SI{50000}{\elementarycharge\per\angstrom\squared}. \stemgym{} provides:
\begin{enumerate}
    \item \textbf{15 annotated HAADF-STEM worlds} generated via PRISM multislice simulation for five materials spanning four crystal structure families, at three difficulty levels.
    \item \textbf{Four characterisation tasks}: defect census (F1), phase mapping (IoU), targeted characterisation (spatial F1), and particle census (composite score).
    \item \textbf{The DEC-AUC metric}: captures both accuracy and convergence speed as a single scalar measuring the dose-accuracy trade-off.
    \item \textbf{33 agent configurations with causal decomposition}: ``equipped'' baselines isolate perception from navigation contributions, plus VLM analysts and three RL baselines.
\end{enumerate}
\section{Related Work}
\label{sec:related}

In autonomous microscopy, Bayesian optimisation has been applied to guide STEM acquisition toward relevant regions~\citep{kalinin2015big, ziatdinov2022hypothesis}, and sparse scanning strategies have shown that much of the specimen need not be irradiated~\citep{stevens2018subsampled}.
Closer to closed-loop autonomous STEM, prior work has demonstrated active acquisition driven by deep-kernel-learning surrogates of the analyst~\citep{roccapriore2022autonomous} and material-specific active-learning loops~\citep{ziatdinov2022bayesian}; both feed analyst output back into navigation, contrasting with the open-loop navigators evaluated here.
More broadly, machine learning is increasingly integrated into STEM workflows for real-time analysis and automated orchestration~\citep{kalinin2023ml4stem, kalinin2021automated}, and adjacent fields face analogous challenges in autonomous synchrotrons~\citep{noack2021gaussian}, scanning probe microscopy~\citep{vasudevan2021autonomous}, and active materials discovery more broadly~\citep{lookman2019active}, with self-driving laboratories~\citep{abolhasani2023selfdriving} and autonomous scientific experimentation~\citep{boiko2023autonomous} demonstrating the broader trend toward closed-loop instrument control.

Benchmarks for autonomous microscopy remain scarce, and microscopy has been argued to lack reproducible benchmarks more broadly~\citep{spurgeon2021towards}; the Microscopy Hackathon~\citep{pratiush2025michackathon} identified this gap as a primary bottleneck for ML adoption. CryoBench recently addressed heterogeneity reconstruction in cryo-EM; no equivalent exists for STEM. MicroBench~\citep{lozano2024microbench} evaluates vision-language models on microscopy image understanding, but targets classification rather than sequential acquisition, and broader scientific multimodal evaluation is anchored by MMMU~\citep{yue2024mmmu}. State-of-the-art atom segmentation networks have been shown to perform poorly outside their training distribution~\citep{wei2023benchmark}, evidence that analyst robustness, not only navigation strategy, deserves systematic measurement.

Statistical methodology for paired benchmark comparisons across agents and tasks follows established practice~\citep{demsar2006statistical}. The materials motivating dose-efficient STEM each present distinct characterisation challenges: perovskite oxides~\citep{borisevich2010mapping}, semiconductor heterostructures~\citep{muller2008atomic}, and catalytic nanoparticles~\citep{nellist2004direct}. These efforts target individual components of the microscopy pipeline; \stemgym{} evaluates the full acquisition loop end-to-end, from navigation through analysis to task scoring, under a unified dose budget.

\section{Method}
\label{sec:method}

\subsection{Problem Formulation}
\label{sec:problem}

We formalize \stemgym{} as \emph{active perception under a hard dose budget}, with performance measured by an anytime dose-efficiency integral. The agent must decide \emph{where} to look and \emph{how} to interpret what it sees, subject to a fixed electron-dose budget that is consumed irreversibly by each acquisition. Symbol conventions used throughout are summarised in SI~\ref{app:notation}.

\noindent\textbf{Environment.}
\stemgym{} is a budget-constrained sequential decision problem
\begin{equation}
\mathcal{M} \;=\; \langle\,\mathcal{S},\,\mathcal{A},\,\mathcal{O},\,T,\,c,\,B,\,H\,\rangle,
\label{eq:mdp-tuple}
\end{equation}
which we describe component by component. The hidden state $\mathcal{S}$ encodes the tile grid and latent material labels of the specimen and is fixed for the duration of an episode. The action space is $\mathcal{A} = \{\textsc{Nav}(x,y),\,\textsc{Acq}(z,d),\,\textsc{Decl},\,\textsc{Done}\}$, covering probe navigation, tile acquisition, finding declaration, and episode termination. The observation space $\mathcal{O}$ comprises a Poisson-noisy $128\!\times\!128$ HAADF tile together with the auxiliary scalars. Stage dynamics $T$ are deterministic. Each episode runs under a dose budget $B = \SI{5000}{\elementarycharge\per\angstrom\squared}$ and a horizon of $H = 200$ steps. The acquisition cost function is
\begin{equation}
c(a) \;=\; d_0\,d\,z^2 \cdot \mathbb{1}[a = \textsc{Acq}],
\label{eq:cost}
\end{equation}
where $d_0 = \SI{100}{\elementarycharge\per\angstrom\squared}$ is a fixed reference dose constant that sets the physical scale, $d$ is the requested dose multiplier, and $z$ is the zoom factor; only \textsc{Acq} actions consume budget. The agent receives no per-step reward; the task score $\rho$ enters only through the offline DEC-AUC integral.

We do not track an explicit belief over $\mathcal{S}$; instead we operate directly on the tile history $h_t = (o_1,\dots,o_t)$ produced by the agent's actions. A decoupled \emph{analyst} $g$ is a function from tile histories to task predictions,
\begin{equation}
g: \mathcal{O}^{\le H} \to \hat{\mathcal{Y}},
\label{eq:analyst-spec}
\end{equation}
implemented as either a U-Net plus CNN ensemble (trained), an intensity-thresholding heuristic (fixed $\pm 1.5$ z-score on tile peak intensities), or a vision--language model. The instantaneous task score is
\begin{equation}
\rho(h_t) \;=\; \mathrm{score}\bigl(g(h_t),\,y^\star\bigr) \;\in\; [0,1],
\label{eq:rho}
\end{equation}
with $y^\star$ the ground-truth specimen annotation. After the agent issues \textsc{Done} or exhausts $B$, $\rho$ is held constant at its final value through dose fraction $f = 1$, so DEC-AUC is well defined for early-terminating policies.

\noindent\textbf{Dose-efficiency curve.}
Let $f_t = B^{-1}\sum_{\tau\le t} c(a_\tau)$ denote the consumed dose fraction and $\tau(f) = \min\{t : f_t \ge f\}$. Adopting an anytime-performance view~\citep{zilberstein1996anytime} and the area-under-learning-curve metric~\citep{guyon2011active}, the dose-efficiency curve is $s(f;\pi,g) = \mathbb{E}_\pi\!\left[\rho(h_{\tau(f)})\right]$, and its area is
\begin{equation}
\mathrm{DEC\text{-}AUC}(\pi,g) \;=\; \int_0^1 s(f;\pi,g)\,df \;\approx\; \sum_{k=1}^{K-1}\frac{s_{k+1}+s_k}{2}\,(f_{k+1}-f_k),
\label{eq:decauc}
\end{equation}
with $K{=}50$ uniformly spaced checkpoints at $f_k = (k-1)/(K-1)$, so $f_1 = 0$ and $f_K = 1$. Score values between checkpoints are linearly interpolated.

The benchmark seeks policy--analyst pairs $(\pi,g)$ that jointly maximize dose efficiency:
\begin{equation}
(\pi^\star,g^\star) \;=\; \arg\max_{\pi,\,g}\; \mathbb{E}_\pi\!\left[\mathrm{DEC\text{-}AUC}(\pi,g)\right].
\label{eq:objective}
\end{equation}
The dose budget $B$ enters only through the cost function $c$ and the support of $f \in [0, 1]$ in Eq.~\ref{eq:decauc}; no separate constraint is needed. This formulation isolates the two levers a real microscopist controls, namely where to look ($\pi$) and how to interpret what was seen ($g$), and rewards both final accuracy and rapid convergence along the dose axis. In all experiments below $g$ is held fixed per agent (the trained analyst, a heuristic, or a VLM); the optimisation is therefore primarily over $\pi$.

\subsection{Environment}
\label{sec:env}

\stemgym{} implements the Gymnasium \texttt{Env} interface with a dictionary action--observation space mirroring a real STEM session. The observation $o_t \sim Z(\cdot\mid s_t,a_t)$ comprises (i)~the acquired $128{\times}128$ HAADF tile, (ii)~a $64{\times}64$ low-resolution overview crop, (iii)~current position, (iv)~remaining dose fraction, (v)~visited-tile mask, and (vi)~step fraction consumed. Episodes terminate at \textsc{Done} or upon budget exhaustion. World files follow the HDF5 layout described in SI~\ref{app:hdf5}.

\subsection{World Generation}
\label{sec:worlds}

Worlds are generated via PRISM multislice simulation~\citep{ophus2017fast} using \texttt{abTEM}~\citep{madsen2021abtem}, tiled into $40{\times}40$ grids (1{,}600 tiles for \ce{SrTiO3}, \ce{BaTiO3}, Pt) or $20{\times}20$ grids (400 tiles for SiGe, GaN). Five material systems span four crystal structure families: \ce{SrTiO3} (cubic perovskite, grain-boundary vacancies), \ce{BaTiO3} (ferroelectric perovskite, phase-boundary vacancies), Si/Ge (diamond cubic, compositional gradient), GaN (wurtzite, InGaN quantum-well substitutions), and Pt nanoparticles on amorphous carbon (aperiodic, three morphologies). All simulations use \SI{200}{\kilo\electronvolt} beam energy, \SI{21}{\milli\radian} convergence, and \SIrange{68}{200}{\milli\radian} HAADF detector angles with literature Debye--Waller factors (SI~\ref{app:simulation}). Each material is generated at three difficulty levels (Table~\ref{tab:difficulty}).

\begin{table}[t]
\centering
\caption{Canonical difficulty parameters; per-material adjustments apply where the underlying physics motivates them (SI~\ref{sec:si:difficulty}). The ``Intended Dose'' column refers to the simulation-time Poisson-noise level that distinguishes the three difficulty tiers, not the runtime acquisition budget.}
\label{tab:difficulty}
\begin{tabular}{lccc}
\toprule
\textbf{Difficulty} & \textbf{Vacancy Rate} & \textbf{Phonon Configs} & \textbf{Intended Dose (\si{\elementarycharge\per\angstrom\squared})} \\
\midrule
Easy   & 5\% & 4  & $10^4$ \\
Medium & 3\% & 8  & $5 \times 10^3$ \\
Hard   & 1\% & 16 & $10^3$ \\
\bottomrule
\end{tabular}
\end{table}

\subsection{Tasks}
\label{sec:tasks}

Each task instantiates the score function $\mathrm{score}(g(b_t), y^\star) \in [0,1]$ of Eq.~\ref{eq:rho} (higher is better). For all except phase mapping, ground truth is filtered to explored tiles.

Defect census uses macro-averaged F1 on defect-type counts, with spatial F1 (within \SI{5}{\nano\meter}) substituted when positions are provided and the maximum of the two reported. Phase mapping uses macro-averaged IoU across phases; reporting \texttt{None} on single-phase materials yields a perfect score. Targeted characterisation uses spatial F1 via nearest-neighbour matching for locating specific defects. Particle census combines $0.5 \cdot F_{1,\text{detect}} + 0.3 \cdot S_\text{sizing} + 0.2 \cdot F_{1,\text{morph}}$. On Pt nanoparticle worlds, ${\sim}85\%$ of tiles contain no particle; a default ``no-particle'' prediction therefore scores high $F_{1,\text{detect}}$ by correctly classifying these majority tiles, explaining the heuristic baseline of $0.908$ in Table~\ref{tab:main}.

\subsection{Dose-Efficiency Curve Metric}
\label{sec:metric}

We operationalize Eq.~\ref{eq:decauc} by recording the task score at regular acquisition intervals and interpolating to the $K{=}50$ checkpoint grid. Agent rankings are robust to metric choice (Spearman $\rho \geq 0.85$ vs four alternatives across $17/20$ task--metric pairs (SI~\ref{app:metric_alternatives}); discretisation error is bounded ($0.016$ between $K{=}20$ and $K{=}50$, $<0.001$ between $K{=}50$ and $K{=}100$). As a secondary diagnostic we report \emph{time-to-threshold} (dose fraction at which the score first reaches $0.8$); it is order-discontinuous when no agent crosses the threshold, motivating DEC-AUC as the primary metric.

\subsection{Agents}
\label{sec:agents}

Each agent instantiates either the navigation policy $\pi$, the analyst $g$, or both (Eq.~\ref{eq:objective}). Three na\"{\i}ve baselines provide reference points: Random performs uniform tile sampling with the heuristic analyst, Raster scans systematically left to right with the same heuristic, and GP-UCB uses a Gaussian process~\citep{rasmussen2006gaussian} over tile interestingness with UCB~\citep{srinivas2010gaussian} acquisition and the heuristic analyst.

\noindent\textbf{Equipped baselines (causal decomposition).}
We create equipped variants of each baseline by pairing their navigation $\pi$ with the trained neural analyst $g$. This $2{\times}2$ factorial design (same analyst with different navigation \emph{vs.}\ same navigation with different analyst) enables causal attribution of performance differences to the two factors of Eq.~\ref{eq:objective}.

STEMAgent is a multi-agent system coordinating (1)~a GP-UCB Navigator with dynamically adjusted $\beta$, (2)~a neural Analyst (U-Net ensemble for atom finding, CNN for defect classification, ResNet for phase identification), and (3)~an FSM Planner transitioning through SURVEY$\to$INVESTIGATE$\to$CHARACTERIZE$\to$CENSUS$\to$TERMINATE modes (SI~\ref{app:stemagent}).

Four ablation variants of STEMAgent quantify component contributions: No Planner ($\beta{=}2.0$ constant), No Uncertainty (single model), Rule Planner (alternative thresholds), and LLM Planner (API-based mode decisions). Production-tier VLM agents (Claude Haiku 4.5, GPT-5 mini, Llama 4 Scout, Gemini 2.0 Flash) replace the CNN analyst via a hybrid approach combining traditional atom localisation with VLM-based classification from upscaled ($256{\times}256$) tiles.

Three RL baselines trained via Stable-Baselines3~\citep{raffin2021stable} for 50{,}000 steps each round out the agent set: DQN~\citep{mnih2015human} and PPO~\citep{schulman2017proximal} use a discrete $8{\times}8$ spatial grid $\times$ 3 action types (192 actions), while SAC~\citep{haarnoja2018soft} uses a continuous $\text{Box}(2)$ action space over normalised coordinates.
\section{Experiments}
\label{sec:experiments}

We evaluate all agents at dose budget \SI{5000}{\elementarycharge\per\angstrom\squared} with 10 seeds per configuration (${\sim}8{,}000$ episodes total; per-experiment breakdown, analyst-training splits, and hyperparameters in SI~\ref{app:compute}, \ref{app:simulation}, \ref{app:hyperparams}). Rankings are robust to scoring cadence (Spearman $\rho > 0.95$; Fig.~\ref{fig:analyses_si}a,b in SI~\ref{app:full_results}). For atom-level tasks (defect census, targeted, particle census), ground truth is filtered to agent-explored tiles---applied uniformly to all agents---so agents are not penalised for regions unreached within budget; phase mapping uses the full-field ground truth since phase labels are spatially extensive.

\subsection{Perception Dominance}
\label{sec:perception}

\begin{table}[t]
\centering
\caption{DEC-AUC (mean $\pm$ std) by agent and task. Entries sharing letter $a$ are pairwise statistically indistinguishable (Wilcoxon $p > 0.05$). $^{*}$std$=0$ indicates a deterministic default prediction across seeds. $^{\dagger\dagger}$The $0$ entry is architectural: the crystallography-trained analyst was not designed for morphological specimens, so this reflects specialisation, not detection failure. $^{\ddagger}$PF GP-BO is evaluated on the five easy worlds only (300 episodes); the high phase-mapping mean reflects trivially-correct phase-$0$ predictions on single-phase worlds (\ce{SrTiO3}, Pt) and the implementation is not a faithful reproduction of the published learned-embedding pipeline (SI~\ref{app:kalinin}). $^{\ddagger\ddagger}$Compressed-sensing baseline (random mask + TV reconstruction + analyst), $20\%$ coverage shown as representative; see SI~\ref{app:compressed} for the $10$/$20$/$30\%$ breakdown and the domain-shift caveat (analyst trained on raw HAADF, fed inpainted tiles). STEMAgent ablations in SI~\ref{app:ablation}.}
\label{tab:main}
\begin{tabular}{l@{\hspace{4pt}}c@{\hspace{4pt}}c@{\hspace{4pt}}c@{\hspace{4pt}}c}
\toprule
\textbf{Agent} & \textbf{Defect Census} & \textbf{Phase Mapping} & \textbf{Targeted} & \textbf{Particle Census} \\
\midrule
Random           & $0.028 \pm 0.027$ & $0.000 \pm 0.000$ & $0.000 \pm 0.000$ & $0.553 \pm 0.353$ \\
Raster           & $0.052 \pm 0.038$ & $0.000 \pm 0.000$ & $0.000 \pm 0.000$ & $0.908 \pm 0.000$ \\
GP-UCB           & $0.034 \pm 0.037$ & $0.000 \pm 0.000$ & $0.000 \pm 0.000$ & $0.562 \pm 0.375$ \\
\midrule
Raster+Analyst   & $0.287^{a} \pm 0.129$ & $0.168 \pm 0.023$ & $0.112 \pm 0.033$ & $0.000^{\dagger\dagger} \pm 0.000$ \\
GP-UCB+Analyst   & $0.283^{a} \pm 0.122$ & $0.184 \pm 0.051$ & $0.127 \pm 0.089$ & $0.000^{\dagger\dagger} \pm 0.000$ \\
STEMAgent        & $0.278^{a} \pm 0.120$ & $0.185 \pm 0.033$ & $0.134 \pm 0.085$ & $0.000^{\dagger\dagger} \pm 0.000$ \\
\midrule
PF GP-BO\textsuperscript{$\dagger$}          & $0.013 \pm 0.006$ & $0.363 \pm 0.000$\textsuperscript{$\ddagger$} & $0.000 \pm 0.000$ & ---             \\
PF GP-BO+Analyst\textsuperscript{$\dagger$}  & $0.163 \pm 0.030$ & $0.128 \pm 0.125$ & $0.348 \pm 0.048$ & ---             \\
\midrule
CS-20\%+Analyst\textsuperscript{$\ddagger\ddagger$} & $0.175 \pm 0.055$ & $0.305 \pm 0.185$ & --- & --- \\
\midrule
DQN              & $0.112 \pm 0.089$ & $0.000 \pm 0.000$ & $0.000 \pm 0.000$ & $0.481 \pm 0.388$ \\
DQN+Analyst      & $0.121 \pm 0.089$ & $0.149 \pm 0.081$ & $0.205 \pm 0.150$ & $0.000 \pm 0.000$ \\
PPO              & $0.002 \pm 0.001$ & $0.000 \pm 0.000$ & $0.000 \pm 0.000$ & $0.010 \pm 0.000$ \\
PPO+Analyst      & $0.001 \pm 0.002$ & $0.000 \pm 0.000$ & $0.009 \pm 0.002$ & $0.000 \pm 0.000$ \\
SAC              & $0.041 \pm 0.049$ & $0.000 \pm 0.000$ & $0.000 \pm 0.000$ & $0.690 \pm 0.378$ \\
SAC+Analyst      & $0.259 \pm 0.139$ & $0.184 \pm 0.027$ & $0.120 \pm 0.096$ & $0.000 \pm 0.000$ \\
\bottomrule
\end{tabular}
\end{table}

Table~\ref{tab:main} reports DEC-AUC across all agent-task combinations. On defect census, equipped agents outperform na\"{\i}ve baselines by roughly $5{.}5\times$ (Raster+Analyst $0.287$ vs Raster $0.052$; paired Wilcoxon $p < 0.001$). Within the equipped group, however, the three navigators are statistically indistinguishable at the tested coverage regime: their DEC-AUC values fall within ${\sim}0.01$ of each other, pairwise Wilcoxon yields $p > 0.05$, bootstrap $95\%$ CIs overlap, and TOST formally supports equivalence at $\Delta = 0.05$ DEC-AUC under two one-sided Wilcoxon tests (entries marked $^{a}$ in Table~\ref{tab:main}; full statistics in SI~\ref{app:statistics}). The within-equipped numerical ordering is therefore not load-bearing; what is supported is that adding adaptive Bayesian navigation or FSM-based planning on top of a trained analyst yields no statistically significant gain over a raster scan with the same analyst. All three navigators are open-loop with respect to analyst output; closed-loop variants that condition acquisition on analyst predictions are not evaluated here.

Once a capable analyst is supplied, per-episode variance ($\sigma \approx 0.12$ for all three equipped agents) is dominated by cross-world heterogeneity rather than agent stochasticity. The effect is sharpest at the per-world level: on \ce{SrTiO3}, three navigators with markedly different exploration trajectories (Random+CNN $0.301$, STEMAgent $0.377$, GP-UCB+CNN $0.394$) cluster within ${\sim}0.09$ DEC-AUC, so distinct \emph{spatial} strategies converge to nearly the same dose-efficiency outcome (trajectories in SI~\ref{app:exploration}). This reinforces that navigation is not the load-bearing factor.

As an external check, the PF GP-BO baseline is competitive with our in-house equipped agents within perovskites but degrades to a ${\sim}14\times$ gap on non-perovskite defect census (SI~\ref{app:kalinin}). The same asymmetry on an independently-implemented prior-art method reinforces that cross-family generalisation, not navigation strategy, is the open problem.

On particle census, the perception ranking inverts: na\"{\i}ve Raster achieves $0.908$ via the no-particle prior (\S\ref{sec:tasks}), while \emph{every} equipped agent scores $0.000$. The crystallography-oriented CNN classifies nanoparticle tiles as anomalous, demonstrating that a strong analyst for one task class can be counterproductive for another. Pt was additionally held out from analyst training (SI~\ref{app:simulation}), so the $0.000$ score reflects both held-out generalisation and architectural specialisation, two effects this benchmark does not separate.

\subsection{Generalisation}
\label{sec:generalisation}

\begin{figure}[t]
\centering
\begin{tikzpicture}
\node[inner sep=4pt, fill=gray!6, rounded corners=3pt, draw=gray!25, line width=0.4pt] {%
\includegraphics[width=0.95\textwidth]{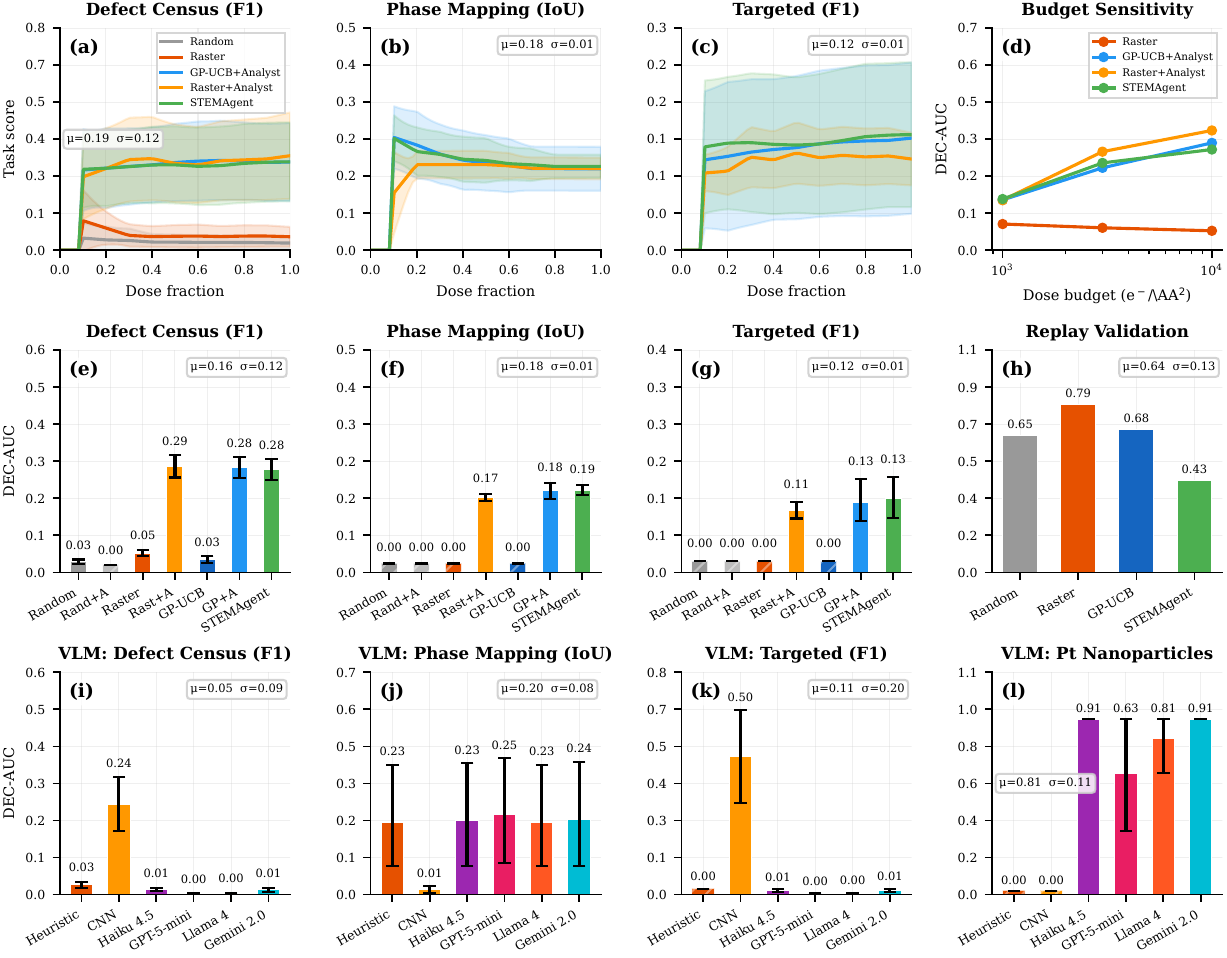}%
};
\end{tikzpicture}
\captionsetup{justification=raggedright,singlelinecheck=false}
\caption{Benchmark results.
\textbf{(a)--(c)}~DEC curves: equipped baselines separate sharply from na\"{\i}ve agents.
\textbf{(d)}~Budget sensitivity: rankings stable across \SI{1000}{} to \SI{10000}{\elementarycharge\per\angstrom\squared}.
\textbf{(e)--(g)}~Bottleneck decomposition (bootstrap 95\% CIs): large perception gap, flat navigation trend.
\textbf{(h)}~Replay validation: Raster (0.792) $>$ GP-UCB (0.676) $>$ Random (0.646) $>$ STEMAgent (0.434).
\textbf{(i)--(k)}~VLM comparison: CNN dominates VLMs on crystallographic worlds (${\sim}13\times$ gap).
\textbf{(l)}~VLM inversion on Pt nanoparticles.}
\label{fig:hero}
\end{figure}

\noindent\textbf{Budget sensitivity.}
Across a $50\times$ dose-budget sweep (\SI{1000}{} to \SI{50000}{\elementarycharge\per\angstrom\squared}, reaching ${\sim}31\%$ coverage at the upper end), the within-equipped ranking is preserved with Raster+Analyst leading; its margin over GP-UCB+Analyst grows from ${\sim}0.005$ at the baseline budget to ${\sim}0.040$. SiGe stays at $0.05$--$0.08$ across all budgets (Figure~\ref{fig:hero}d), the cleanest illustration that material-specific analyst limits dominate over measurement strategy (SI~\ref{app:full_results}).

\noindent\textbf{Difficulty scaling.}
On \ce{BaTiO3} across three difficulty levels (vacancy rate $4\%\to3\%\to1.5\%$, $1\to3$ wavy phase boundaries, $4\to16$ frozen-phonon configurations), equipped agents stay within a narrow $0.293$--$0.359$ band on defect census while na\"{\i}ve baselines remain below $0.10$ and na\"{\i}ve phase mapping is exactly zero at every level (SI~\ref{sec:si:difficulty}).

\noindent\textbf{Compressed sensing.}
A sparse random mask with total-variation reconstruction~\citep{candes2006robust,chambolle2004algorithm} feeding the trained analyst reaches $0.149$--$0.175$ DEC-AUC across $10$/$20$/$30\%$ mask coverage on the same crystalline worlds (Table~\ref{tab:main}). The comparison is diagnostic given the domain shift from raw HAADF to TV-inpainted tiles (SI~\ref{app:compressed}).

\noindent\textbf{Replay validation.}
A held-out synthetic specimen (the \texttt{replay\_world} HDF5 file in our release) preserves the equipped-vs-na\"{\i}ve family ordering (Raster $>$ GP-UCB $>$ Random; Figure~\ref{fig:hero}h), with STEMAgent ranking below all three on this specimen (SI~\ref{app:datasheet}).

\noindent\textbf{Sim-to-real anchor.}
On five real Sm-doped \ce{BiFeO3} acquisitions (Sm $\in \{0, 7, 10, 13, 20\}\%$)~\citep{ghosh2021bfo}, surfaced via the 2024 Mic-hackathon team~14 notebook, Raster+Analyst attains $0.066$ and $0.088$ DEC-AUC on the Sm $0\%$ and $13\%$ compositions and degenerates to $0.000$ on the other three (zero-shot transfer of a perovskite-trained analyst to a chemically-different perovskite). The heuristic raster baseline scores $0.000$ on every composition; the equipped-vs-na\"{\i}ve direction is preserved or tied on all five and never inverts (SI~\ref{app:realdata}).

\subsection{VLM Comparison}
\label{sec:vlm}

Four production-tier VLMs (Claude Haiku 4.5, GPT-5 mini, Llama 4 Scout, Gemini 2.0 Flash) are paired with Raster and GP-UCB navigation and compared against heuristic and CNN baselines on three tasks across the five easy worlds (760 effective episodes after GPT-5 mini was excluded for ${\sim}100\%$ retry-exhausted empty-content failures). To test whether the perception gap is a tier-of-model artefact, we additionally ran a frontier-flagship sanity check with Claude Opus 4.7 on defect census ($5$ worlds $\times$ $3$ seeds, raster navigation, no parse failures): mean DEC-AUC $0.016$ on the four crystalline worlds, sitting inside the production-tier band and ${\sim}15\times$ below Raster+Analyst. Methodology, prompts, retry harness, and per-model diagnostics are in SI~\ref{app:vlm_details}--\ref{app:vlm_diagnostics}.

On crystallographic worlds (\ce{SrTiO3}, \ce{BaTiO3}, SiGe, GaN), the trained CNN dominates: Raster+Analyst averages $0.247$ DEC-AUC on defect census versus the best VLM (Raster+Gemini) at $0.019$, a ${\sim}13\times$ gap, with Raster+Claude at $0.006$ and Raster+Llama at $0.019$. On Pt nanoparticles the pattern inverts: Raster+Claude matches the trivially-correct na\"{\i}ve Raster score (\S\ref{sec:tasks}), Raster+Gemini reaches $0.628$, and Raster+Llama reaches $0.458$, while the CNN scores $0.000$ (the particle-census inversion). The CNN was trained for atomic-column localisation and treats aperiodic features as anomalies, whereas production-tier VLMs encode general visual priors that recognise dispersed bright objects on textured substrates; the architectures are genuinely complementary, and the result motivates hybrid perception that routes to the appropriate analyst per specimen type rather than choosing a single one. This complementarity is symmetric in aggregate: across all five easy worlds, Raster+CNN ($0.198$) and Raster+Claude Haiku ($0.186$) tie within ${\sim}6\%$ overall (SI~\ref{app:vlm}). The crystallographic gap is therefore a within-domain finding rather than a global verdict on perception architecture, and which family ``wins'' depends on the specimen-type weighting of the evaluation suite. On the targeted task the inversion does not appear: the CNN reaches $0.490$ on crystalline worlds while every VLM stays below $0.020$, and all approaches score $0.000$ on Pt. VLMs are useful for coarse scene-level classification but not for the spatially precise localisation that targeted characterisation requires.

The working models had ${\le}3$ parse failures across the full run, so DEC-AUC reflects actual perception limits rather than I/O artefacts. Per-step wall-clock latency for all agent classes is tabulated in SI~\ref{app:compute}.

A $4{\times}4$ cross-material transfer matrix (rows: training material; columns: evaluation material) reveals strong within-perovskite generalisation: \ce{SrTiO3}-trained models reach $0.723$ on \ce{BaTiO3} and \ce{BaTiO3}-trained models reach $0.830$ on \ce{SrTiO3}. Cross-family transfer is poor: SiGe in-domain $0.039$, GaN $0.018$ (Pt excluded; morphological labels are incompatible). The same asymmetry appears in the PF GP-BO baseline, corroborating that the gap is a property of cross-family transfer rather than our particular analyst. Full matrix in SI~\ref{app:full_results}; PF GP-BO breakdown in SI~\ref{app:kalinin}.

Reinforcement learning offers a contrasting baseline. Three RL agents (DQN, PPO, SAC) trained for $50{,}000$ steps on the easy worlds score near random without trained perception (DQN $0.112$, SAC $0.041$, PPO $0.002$). With the analyst, SAC+Analyst ($0.259$) approaches Raster+Analyst while PPO+Analyst ($0.001$) collapses entirely: algorithm class matters when perception is available, and the sparse terminal reward in a $200$-step horizon penalises on-policy methods most. Acquisition-function comparison (UCB vs.\ EI vs.\ TS) and STEMAgent ablations are statistically indistinguishable (max spread $< 0.024$). Full RL configurations and ablation tables are in SI~\ref{app:rl} and~\ref{app:ablation}.
\section{Limitations}
\label{sec:limitations}

\noindent\textbf{Validation scope.}
\stemgym{} worlds use PRISM multislice with frozen-phonon Debye--Waller factors and Poisson detector noise; they omit inelastic scattering, specimen drift, surface contamination, beam-induced damage accumulation, and detector non-linearity. The five-composition Sm-doped \ce{BiFeO3} sanity check (\S\ref{sec:generalisation}) anchors the equipped-vs-na\"{\i}ve direction on real HAADF-STEM acquisitions but does not constitute multi-instrument cross-material validation. The trained analyst is fitted on four easy-difficulty worlds (\ce{SrTiO3}, \ce{BaTiO3}, SiGe, GaN) and evaluated on rollouts that traverse the full tile grid of every evaluated world, so equipped DEC-AUC reflects \emph{near-distribution} generalisation within seen material families; strict cross-family transfer is reported separately (Pt-held-out particle census; SI~\ref{app:full_results}).

\noindent\textbf{Open-loop navigation caveat.}
All evaluated navigators are open-loop with respect to analyst output: acquisition decisions do not condition on the analyst's predictions. Closed-loop variants such as analyst-uncertainty-guided GP-UCB would be the natural next test of the perception-dominance claim. Compressed sensing addresses \emph{how} to reconstruct rather than \emph{where} to look; the CS+Analyst row is diagnostic given the raw-HAADF-vs-TV-inpainted domain shift, and a retrained analyst on inpainted tiles would isolate the two effects (SI~\ref{app:compressed}). Navigation is modelled as dose-free and latency-free; real-instrument repositioning costs would only reduce the advantage of adaptive strategies over raster scanning, reinforcing the second-order role of navigation (a movement-penalised extension is sketched in SI~\ref{app:nav_cost}).

\noindent\textbf{Coverage and training scale.}
Five material systems span four crystal-structure families (perovskite, diamond cubic, wurtzite, FCC metallic), but three are perovskite or perovskite-adjacent; MXenes, MOFs, 2D chalcogenides, organics, and amorphous solids are not yet included. The three RL agents are trained for $50{,}000$ environment steps, near the lower end of typical RL budgets.

\section{Discussion}
\label{sec:discussion}

The dominant lever for dose efficiency is analysis: open-loop navigation strategies that do not condition on analyst output contribute negligibly once a competent analyst is supplied. Equipping raster scanning with trained perception yields a $5.5\times$ improvement at the baseline budget and remains the primary determinant across the full \SI{1000}{}--\SI{50000}{\elementarycharge\per\angstrom\squared} sweep, with navigation and planning second-order across all difficulty settings and dose budgets. The same direction is preserved on the five-composition Sm-doped \ce{BiFeO3} real-data sanity check: zero-shot transfer of a perovskite-trained analyst to real \ce{BiFeO3} acquisitions never inverts the equipped-vs-na\"{\i}ve direction across a controlled doping gradient (SI~\ref{app:realdata}).

The mechanism behind the flat navigation trend is straightforward: the GP-UCB navigator optimises intensity variance, a proxy that does not predict defect presence, and the FSM planner's $\beta$ transitions barely alter behaviour on a $40{\times}40$ grid where broad exploration visits only ${\sim}50$ of $1{,}600$ tiles. Closing the loop between perception and navigation is therefore the most promising direction for further gains, and the DEC-AUC of $0.287$ on defect census reflects task difficulty in this regime rather than metric saturation: substantial headroom remains for closed-loop navigation and learned-embedding analysts.

Specialisation within the analyst is the next frontier. The particle-census inversion (CNN $0.000$, heuristic $0.908$) and the VLM/CNN complementarity on aperiodic vs.\ crystalline specimens together motivate a learned perception router that dispatches to the appropriate analyst per specimen type. Within-family generalisation is supported by the \ce{SrTiO3}$\leftrightarrow$\ce{BaTiO3} pair ($0.72$--$0.83$ DEC-AUC, corroborated by the PF GP-BO baseline; \S\ref{sec:vlm}), and extending to a third perovskite or to non-perovskite crystal classes (MXenes, MOFs) is a natural follow-up enabled by the HDF5 world format.



\noindent\textbf{Reproducibility Statement.}
All code, data generation scripts, trained model checkpoints, and experiment configurations are publicly available at \url{https://github.com/KurbanIntelligenceLab/STEMGym}, with pre-generated world files on Hugging Face as \textit{stem-gym-benchmark}. Code and data are released under MIT/CC-BY-4.0; extended environment-setup, dependency, and re-run instructions are in SI~\ref{app:reproducibility}.

\bibliographystyle{unsrtnat}
\bibliography{main}


\newpage
\appendix
\section{Notation Summary}
\label{app:notation}

\begin{table}[h]
\centering
\caption{Notation used throughout \S\ref{sec:method}--\S\ref{sec:experiments}.}
\label{tab:notation}
\small
\begin{tabular}{lp{0.62\linewidth}}
\toprule
\textbf{Symbol} & \textbf{Meaning} \\
\midrule
$B$ & Per-episode dose budget in \si{\elementarycharge\per\angstrom\squared} (default $5{,}000$). \\
$H$ & Episode horizon in environment steps (default $200$). \\
$d_0$ & Reference dose constant, $\SI{100}{\elementarycharge\per\angstrom\squared}$. \\
$d$ & Per-acquisition dose multiplier passed in the action; appears as $d$ in Eq.~\ref{eq:cost} and SI~\ref{app:nav_cost}. \\
$c$ & Per-action dose cost (Eq.~\ref{eq:cost}); zero for $\textsc{Nav}$, $\textsc{Decl}$, $\textsc{Done}$. \\
$z$ & Zoom factor (per-acquisition action argument); distinct from atomic number $Z$. \\
$Z$ & Atomic number (Introduction; appears only in $Z^{1.7}$ HAADF-contrast scaling). \\
$f, f_t$ & Consumed dose fraction $f_t = B^{-1}\sum_{\tau\le t} c(a_\tau)$. \\
$\tau(f)$ & DEC hitting time, $\min\{t : f_t \ge f\}$; not to be confused with the SAC Polyak coefficient $\tau_\text{pol}$ in SI~\ref{app:rl}. \\
$K$ & Number of DEC-AUC checkpoints (default $50$). \\
$y^\star$ & Ground-truth task target. \\
$g$ & Analyst, $g: \mathcal{O}^{\le H} \to \hat{\mathcal{Y}}$ (trained ensemble, heuristic, or VLM). \\
$\rho$ & Task score $\rho(h_t) \in [0,1]$. \\
$\beta$ & GP-UCB exploration coefficient; the FSM mode-dependent value is documented in SI~\ref{app:stemagent}. \\
$\sigma$ & Standard deviation of DEC-AUC over seeds. We use $\sigma_G$ for the Gaussian target width in the AtomFinderUNet (SI~\ref{app:simulation}) and $\sigma_\partial$ for grain-boundary sharpness in the world generators when those distinct meanings are needed. \\
\bottomrule
\end{tabular}
\end{table}

\section{HDF5 World Format}
\label{app:hdf5}

\begin{verbatim}
/metadata/                    # Scalar attributes
    pixel_size_nm      float  # Nanometers per pixel
    tile_size_px       int    # Tile edge length (128)
    overlap_px         int    # Overlap between tiles (4)
    grid_shape         (2,)   # (rows, cols) tile grid
    fov_nm             (2,)   # Field of view in nm
/overview              (H, W) float32  # Low-res overview
/tiles/{row}_{col}     (128, 128) float32  # Normalized HAADF
/ground_truth/
    atom_positions     (N, 2) float32  # nm coordinates
    atom_types         (N,)   int32    # 0=pristine, 1=vac, 2=sub
    defect_mask        (N,)   bool     # True if defect
    phase_map          (H, W) int32    # Optional
/valid_region          (H, W) bool
\end{verbatim}

\section{Simulation Parameters}
\label{app:simulation}

All simulations use \SI{200}{\kilo\electronvolt} beam energy, \SI{21}{\milli\radian} convergence semi-angle, and HAADF detector angles of \SIrange{68}{200}{\milli\radian}.
Frozen-phonon thermal diffuse scattering uses literature RMS displacements:
Sr \SI{0.084}{\angstrom}, Ti \SI{0.060}{\angstrom}, O \SI{0.085}{\angstrom} for \ce{SrTiO3}~\citep{abramov1995distribution};
Ba \SI{0.080}{\angstrom}, Ti \SI{0.080}{\angstrom}, O \SI{0.090}{\angstrom} for \ce{BaTiO3}~\citep{kwei1993structures} (elevated Ti displacement from order--disorder ferroelectric transition);
Si \SI{0.078}{\angstrom}, Ge \SI{0.085}{\angstrom} for SiGe~\citep{peng1996electron};
Ga \SI{0.059}{\angstrom}, N \SI{0.063}{\angstrom}, In \SI{0.065}{\angstrom} for GaN~\citep{schowalter2009temperature};
Pt \SI{0.070}{\angstrom}, C \SI{0.080}{\angstrom} for Pt/C~\citep{peng1996electron}.
PRISM interpolation factor is 2, with \SI{0.05}{\angstrom} real-space sampling and \SI{1.0}{\angstrom} slice thickness.
Large fields of view are simulated as \SI{7}{\nano\meter} sub-tiles with \SI{1}{\nano\meter} overlap, blended via linear ramps.

\begin{figure}[h]
\centering
\begin{tikzpicture}
\node[inner sep=4pt, fill=gray!6, rounded corners=3pt, draw=gray!25, line width=0.4pt] {%
\includegraphics[width=0.95\textwidth]{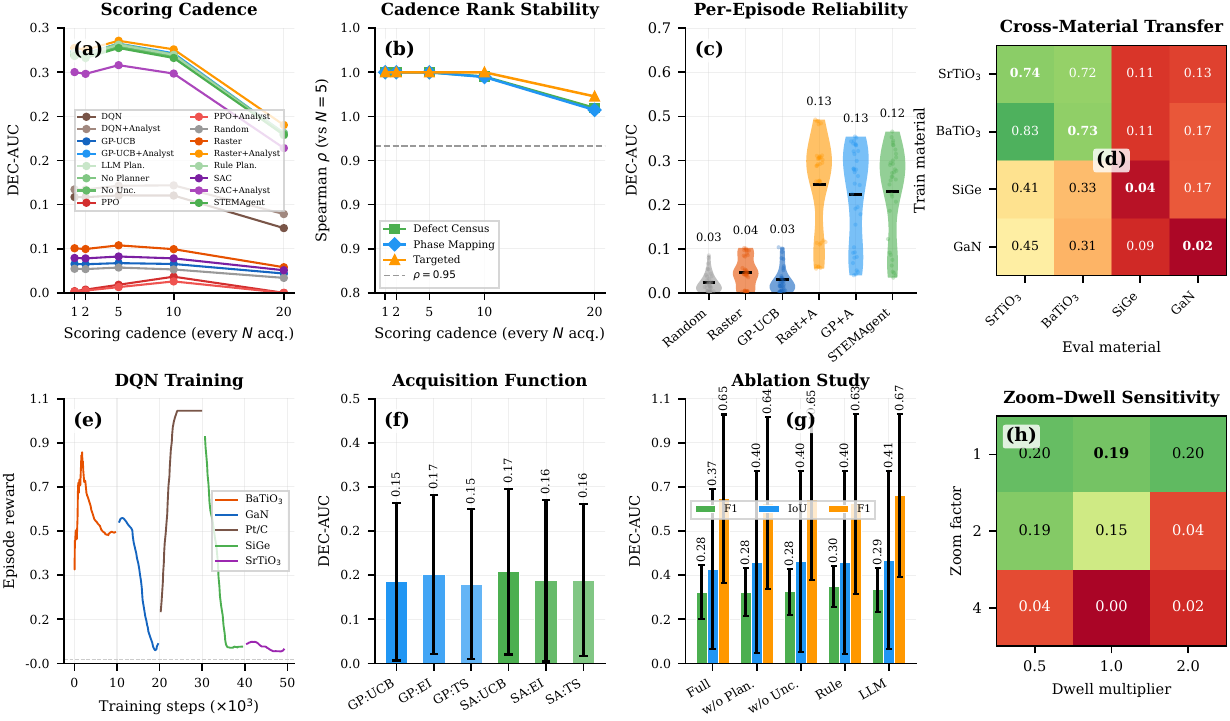}%
};
\end{tikzpicture}
\captionsetup{justification=raggedright,singlelinecheck=false}
\caption{Quantitative analyses (relocated from main text for space).
\textbf{(a)}~Scoring cadence stability.
\textbf{(b)}~Cadence rank stability (Spearman $\rho > 0.95$).
\textbf{(c)}~Per-episode reliability distributions.
\textbf{(d)}~Cross-material transfer matrix (rows: training material; columns: evaluation material).
\textbf{(e)}~DQN training curve.
\textbf{(f)}~Acquisition function comparison.
\textbf{(g)}~STEMAgent ablation.
\textbf{(h)}~Zoom--dwell sensitivity.}
\label{fig:analyses_si}
\end{figure}

\paragraph{Analyst Training.}
The ensemble is trained on tiles from four easy-difficulty worlds (\ce{SrTiO3}, \ce{BaTiO3}, SiGe, GaN; ${\sim}$3{,}000 tiles total), with Pt excluded as held-out material.
Within each of the four training worlds, tiles are split 80/20 (seed 42) at the \emph{tile} level. There is no held-out evaluation \emph{world} for the four training materials; agent rollouts at evaluation time traverse the full grid of all evaluated worlds, including both training-split and validation-split tiles. Equipped DEC-AUC therefore reflects \emph{near-distribution} generalisation within seen material families. Strict cross-family generalisation is reported separately in the cross-material transfer matrix (Fig.~\ref{fig:analyses_si}d, SI~\ref{app:full_results}) and the Pt-held-out particle census.
AtomFinder targets are Gaussian probability maps ($\sigma{=}2$~px) with BCE loss.
DefectClassifier uses $32{\times}32$ patches with class-balanced subsampling (max 50{,}000 patches) and weighted CE.
PhaseIdentifier trains only on multi-phase worlds (\ce{BaTiO3}, GaN).
Augmentation: random 90\textdegree{} rotations, horizontal flips ($p{=}0.5$), additive Gaussian noise ($\sigma \sim \mathcal{U}[0, 0.05]$), and gamma jitter ($\gamma \sim \mathcal{U}[0.8, 1.2]$).
All models use Adam ($\text{lr}=10^{-3}$) with early stopping (patience 7).

\section{DEC-AUC Stability Across Alternative Metrics}
\label{app:metric_alternatives}

DEC-AUC summarises the full dose-score trajectory as a single scalar via trapezoidal integration. To check whether agent rankings are robust to this choice, we recomputed five alternative scoring metrics from the saved per-episode \texttt{dec\_dose\_fractions} and \texttt{dec\_scores} arrays and compared rankings across all reported configurations: (i) \emph{final accuracy} $s(f{=}1)$, the simplest possible metric; (ii) \emph{time-to-threshold} $\min\{f : s(f) \geq 0.8\}$ with $1.0$ for never-reached; (iii) \emph{early-weighted AUC} $\int_0^1 s(f)\cdot 2(1-f)\,df$ privileging front-loaded acquisition; (iv) \emph{late-weighted AUC} $\int_0^1 s(f)\cdot 2f\,df$ privileging asymptotic accuracy; (v) \emph{half-budget AUC} $\frac{1}{0.5}\int_0^{0.5} s(f)\,df$ truncating to the first half of dose. For each task we ranked agents under each metric and computed Spearman $\rho$ between DEC-AUC and each alternative; \texttt{time-to-threshold} is internally negated so the correlation measures ranking agreement rather than monotone direction. The analysis covers $105$ unique (agent, task) pairs across the four tasks (defect census $32$ agents, phase mapping $32$, targeted $26$, particle census $15$).

\begin{table}[h]
\centering
\caption{Spearman $\rho$ between DEC-AUC ranking and alternative-metric rankings, per task. Values close to 1 indicate the agent ordering is preserved under the alternative metric, supporting DEC-AUC's stability as a benchmark choice. \texttt{time-to-threshold} is internally negated (lower-is-better) before ranking so that the correlation reflects agreement, not direction.}
\label{tab:metric_alternatives}
\begin{tabular}{lcccc}
\toprule
\textbf{Alternative metric} & \textbf{Defect Census} & \textbf{Phase Mapping} & \textbf{Targeted} & \textbf{Particle Census} \\
\midrule
Final accuracy & $0.880$ & $0.900$ & $0.931$ & $0.917$ \\
Time-to-threshold & $0.097$ & $0.861$ & $0.641$ & $0.829$ \\
Early-weighted AUC & $0.994$ & $0.973$ & $0.977$ & $0.872$ \\
Late-weighted AUC & $0.994$ & $0.994$ & $0.998$ & $0.986$ \\
Half-budget AUC & $0.984$ & $0.961$ & $0.974$ & $0.854$ \\
\bottomrule
\end{tabular}
\end{table}

\paragraph{Continuous AUC variants are essentially equivalent.}
The three trajectory-integral alternatives (early-weighted, late-weighted, half-budget) all produce Spearman $\rho \geq 0.85$ against DEC-AUC across every task, with late-weighted AUC almost indistinguishable from DEC-AUC ($\rho \geq 0.986$ on all four tasks). The latter is a consequence of how DEC curves actually look in our setting: most agents spend the first half of their dose budget at low scores and only accumulate accuracy near the end, so weighting late fractions more heavily preserves the same ranking. Front-loading via the early-weighted alternative perturbs particle census slightly ($\rho = 0.872$) because the heuristic raster baseline that wins that task accumulates score gradually rather than late, but it remains in close agreement on the three crystallographic tasks ($\rho \geq 0.97$).

\paragraph{Final accuracy is robust but loses dose efficiency.}
Final accuracy ranks agents the same way as DEC-AUC at $\rho = 0.88$--$0.93$ depending on task. The disagreements concentrate on agents that converge slowly: an agent that reaches the same final score as another via a longer dose-fraction path is rated identically by final accuracy but lower by DEC-AUC. This is precisely the dose-efficiency information DEC-AUC is designed to retain.

\paragraph{Time-to-threshold collapses on the hardest tasks.}
The most striking disagreement is \texttt{time-to-threshold} on defect census ($\rho = 0.097$). The threshold of $0.8$ is rarely crossed on this task: at the main-table dose budget of $5{,}000$~e/\AA$^2$ ($\sim$3\% coverage), no agent's mean DEC curve reaches $0.8$ on the crystalline easy worlds. Because the metric maps every never-reaching agent to the same value ($1.0$ by convention), the resulting ranking degenerates to a near-constant vector and Spearman correlation with any other metric becomes noise. On phase mapping and targeted the threshold is reached more often (single-phase worlds trivially saturate phase mapping; targeted has fewer false positives) and the correlation rises into the $0.64$--$0.86$ range; on particle census the heuristic raster baseline reliably crosses $0.8$ and the correlation reaches $0.829$. The defect-census collapse is itself the order-discontinuity signature that motivates DEC-AUC: a smooth trajectory integral preserves ranking information even when no agent crosses a threshold, whereas time-to-threshold loses all signal in that regime.

\paragraph{Practical implication.}
Across the four tasks and five alternatives, $19$ of $20$ Spearman $\rho$ values lie above $0.64$ and $17$ lie above $0.85$; the single outlier (time-to-threshold on defect census) is a known degenerate case explained above. DEC-AUC is therefore not the only sensible scalar summary for our setting, but it is the most robust: it ranks agents the same way as the smoother alternatives and avoids the threshold-degeneracy of time-to-threshold. The full per-(agent, task) metric values and per-task disagreement records are released alongside the benchmark in the supplementary archive.

\paragraph{Discretisation error.}
Re-interpolating saved \texttt{dec\_dose\_fractions} and \texttt{dec\_scores} arrays from $n = 4340$ episodes to $K \in \{20, 50, 100\}$ uniformly-spaced checkpoints and recomputing DEC-AUC by trapezoidal integration produces the following maximum absolute deltas: $\max_{K=20 \leftrightarrow 50} |\Delta| = 0.0161$, $\max_{K=50 \leftrightarrow 100} |\Delta| = 0.0001$, $\max_{K=20 \leftrightarrow 100} |\Delta| = 0.0161$. The 99th-percentile delta across all pairs is $0.0161$. Discretisation error at the default $K=50$ is therefore well below the inter-agent gaps (${\sim}0.005$ vs.\ ${\sim}0.235$ for the Raster vs.\ Raster+Analyst perception gap), validating the trapezoidal integration choice.

\paragraph{F1 construction sensitivity.}
Defect census reports $\max(\text{count-F1}, \text{spatial-F1})$ when agents provide positions (\S\ref{sec:tasks}). A controlled sensitivity analysis that reports rankings under count-F1 only, spatial-F1 only, and the max requires re-running every episode with raw findings serialisation enabled (current logs save only the scalar DEC-AUC). We treat this as a low-cost extension for the next benchmark release; the rankings reported here use the max construction.

\section{STEMAgent Architecture Details}
\label{app:stemagent}

\begin{figure}[h]
\centering
\begin{tikzpicture}
\node[inner sep=4pt, fill=gray!6, rounded corners=3pt, draw=gray!25, line width=0.4pt] {%
\begin{minipage}{0.95\textwidth}
\centering
\begin{minipage}[t]{0.55\textwidth}
\centering
\textbf{(a)} Architecture\\[2pt]
\includegraphics[width=\textwidth]{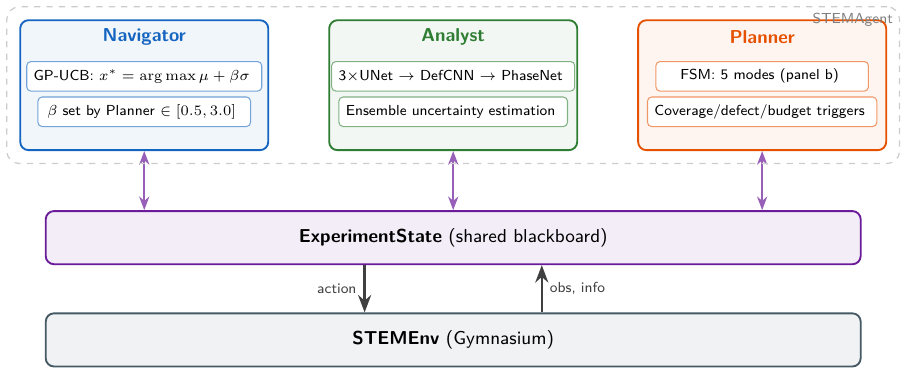}
\end{minipage}\hfill
\begin{minipage}[t]{0.42\textwidth}
\centering
\textbf{(b)} Planner FSM\\[2pt]
\includegraphics[width=\textwidth]{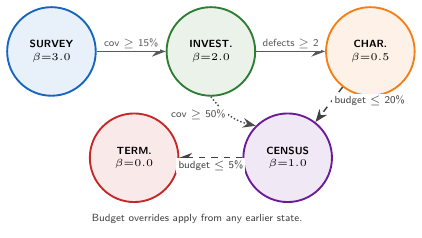}
\end{minipage}
\end{minipage}%
};
\end{tikzpicture}
\caption{STEMAgent design. \textbf{(a)} Three sub-agents communicate via a shared ExperimentState blackboard. \textbf{(b)} Planner FSM: solid arrows show normal transitions; dotted arrow shows coverage fallback; dashed arrows show budget overrides.}
\label{fig:stemagent}
\end{figure}

Figure~\ref{fig:stemagent} shows the full STEMAgent architecture and planner FSM.

\paragraph{Navigator.}
GP-UCB with exploration parameter $\beta$ dynamically set by the Planner (SURVEY: $\beta{=}3.0$; CHARACTERIZE: $\beta{=}0.5$).

\paragraph{Analyst.}
\label{sec:analyst}
Neural ensemble: (1) AtomFinderUNet~\citep{ronneberger2015unet} (3-member ensemble for uncertainty), (2) DefectClassifierCNN, (3) PhaseIdentifierResNet~\citep{he2016deep}.

\paragraph{Heuristic analyst.}
\label{sec:heuristic_analyst}
When no trained ensemble is available, the analyst falls back to a fixed-threshold rule that flags a tile peak intensity as a candidate defect when its $z$-score (against the per-episode running mean and standard deviation of acquired-tile peak intensities) exceeds $\pm 1.5$. Defect type is assigned by intensity sign relative to the local lattice mean: above-mean peaks are tagged as substitutions and below-mean peaks as vacancies. Phase identification under the heuristic is an adaptive intensity median split with a Cohen's $d$ test for bimodality; single-mode tiles return no phase prediction. The threshold is constant across worlds and difficulties and matches the value reported in the difficulty-scaling analysis (\S\ref{sec:si:difficulty}).

\paragraph{Planner.}
FSM with five modes: SURVEY ($\beta{=}3.0$, until 15\% coverage) $\to$ INVESTIGATE ($\beta{=}2.0$, until $\geq 2$ defects) $\to$ CHARACTERIZE ($\beta{=}0.5$) $\to$ CENSUS ($\beta{=}1.0$, $< 20\%$ budget) $\to$ TERMINATE ($< 5\%$ budget).

\section{Hyperparameters}
\label{app:hyperparams}
Table~\ref{tab:agentparams} summarizes the operational hyperparameters of the agent, organized by component. The Navigator (GP-UCB) controls exploration using a Gaussian Process with an exploration parameter $\beta \in [0.5, 3.0]$, starting from 4 random samples and refitting every 3 steps. The Planner (FSM) governs high-level behavior transitions based on coverage, defect detection, and remaining budget thresholds (e.g., 15\% survey coverage, 5\% termination threshold). The Analyst comprises neural models with a 3-member UNet ensemble for localization, a CNN with 3 defect classes, and a ResNet with 2 phase classes, operating under a spatial tolerance of \SI{5}{\nano\meter}. The Environment parameters define acquisition constraints, including $128\times128$~px tiles with 4~px overlap, a fixed dose per acquisition, and a maximum of 200 steps per episode.

Table~\ref{tab:training} details the training configuration of the Analyst models. The AtomFinder (UNet) uses a 3-member ensemble trained with binary cross-entropy loss, batch size 8, and up to 30 epochs, with a target Gaussian spread of $\sigma=2$~px. The DefectClassifier (CNN) operates on $32\times32$~px patches with weighted cross-entropy to address class imbalance, using batch size 32 and up to 50{,}000 patches. The PhaseIdentifier (ResNet) is trained with standard cross-entropy, batch size 8, and 20 epochs. All models share the Adam optimizer with learning rate $10^{-3}$, early stopping with patience of 7 epochs, an 80/20 train-validation split (seed 42), and training data drawn from four material systems (\ce{SrTiO3}, \ce{BaTiO3}, SiGe, GaN).

\begin{table}[h]
\centering
\caption{Agent hyperparameters.}
\label{tab:agentparams}
\begin{tabular}{llr}
\toprule
\textbf{Component} & \textbf{Parameter} & \textbf{Value} \\
\midrule
Navigator (GP-UCB) & Initial random samples & 4 \\
 & GP refit interval & 3 \\
 & $\beta$ range & $[0.5, 3.0]$ \\
\midrule
Planner (FSM) & Survey coverage threshold & 15\% \\
 & Investigate defect threshold & 2 \\
 & Census budget threshold & 20\% \\
 & Terminate budget threshold & 5\% \\
\midrule
Analyst (Neural) & UNet ensemble members & 3 \\
 & CNN defect classes & 3 \\
 & ResNet phase classes & 2 \\
 & Spatial tolerance & \SI{5}{\nano\meter} \\
\midrule
Environment & Tile size & $128 \times 128$~px \\
 & Overlap & 4~px \\
 & Dose per acquire & \SI{100}{\elementarycharge\per\angstrom\squared} \\
 & Max steps & 200 \\
\bottomrule
\end{tabular}
\end{table}

\begin{table}[h]
\centering
\caption{Analyst training hyperparameters.}
\label{tab:training}
\begin{tabular}{llr}
\toprule
\textbf{Model} & \textbf{Parameter} & \textbf{Value} \\
\midrule
AtomFinder (UNet) & Ensemble members & 3 \\
 & Target $\sigma$ & 2~px \\
 & Loss & BCE \\
 & Batch size & 8 \\
 & Max epochs & 30 \\
\midrule
DefectClassifier (CNN) & Patch size & $32 \times 32$~px \\
 & Max patches & 50{,}000 \\
 & Loss & Weighted CE (inv.\ freq.) \\
 & Batch size & 32 \\
 & Max epochs & 30 \\
\midrule
PhaseIdentifier (ResNet) & Loss & CE \\
 & Batch size & 8 \\
 & Max epochs & 20 \\
\midrule
All models & Optimiser & Adam \\
 & Learning rate & $10^{-3}$ \\
 & Early stopping patience & 7 epochs \\
 & Train/val split & 80/20 (seed 42) \\
 & Training worlds & 4 easy (\ce{SrTiO3}, \ce{BaTiO3}, SiGe, GaN) \\
\bottomrule
\end{tabular}
\end{table}

\section{Navigation Cost Model}
\label{app:nav_cost}

The current environment models navigation as dose-free and latency-free.
Real STEM instruments incur time-dependent costs during navigation (stage settling, beam blanking, hysteresis).
Dose cost is $c = d_0 \cdot d \cdot z^2$ where $d_0$ is the base dose, $d$ the dose multiplier (matching Eq.~\ref{eq:cost} in the main text), and $z$ the zoom factor.
Incorporating a movement penalty or time budget is a natural extension that would test whether path planning becomes beneficial when navigation is costly.

\section{Statistical Analysis}
\label{app:statistics}
Table~\ref{tab:statistics} reports bootstrap 95\% confidence intervals for the mean DEC-AUC across agents, computed from 10{,}000 resamples over $N=70$ episodes (10 seeds $\times$ 7 (world, difficulty) cells). The results show that baseline strategies (Random, Raster, GP-UCB) achieve low performance, while adding the Analyst yields a substantial increase in DEC-AUC (e.g., $\approx 0.28$ for Raster+Analyst, GP-UCB+Analyst, and STEMAgent). The Analyst-equipped agents have overlapping confidence intervals; we note that overlapping CIs are not equivalent to non-significance, so the formal claim of comparable performance rests on the paired Wilcoxon test (Table~\ref{tab:wilcoxon}) and the TOST equivalence test below, not on CI overlap alone.

Table~\ref{tab:wilcoxon} presents paired Wilcoxon signed-rank tests on matched episodes to assess statistical significance of performance differences. The comparison between Raster and Raster+Analyst shows a large and statistically significant improvement ($\Delta=+0.235$, $p<0.001$). In contrast, differences among Analyst-based agents (Raster+Analyst, GP-UCB+Analyst, STEMAgent) are small and not statistically significant at $N=70$ ($p>0.05$); the formal equivalence claim is supported by the TOST analysis in Table~\ref{tab:tost}. Ablation tests on STEMAgent (removing uncertainty or planner) also show negligible and non-significant changes, suggesting these components do not materially affect DEC-AUC under the tested conditions.

\begin{table}[t]
\centering
\caption{Bootstrap 95\% confidence intervals on mean DEC-AUC (defect census, 10{,}000 resamples).}
\label{tab:statistics}
\begin{tabular}{lrcc}
\toprule
\textbf{Agent} & \textbf{$N$} & \textbf{Mean DEC-AUC} & \textbf{95\% CI} \\
\midrule
Random & 70 & $0.028$ & $[0.022,\, 0.035]$ \\
Raster & 70 & $0.052$ & $[0.043,\, 0.061]$ \\
GP-UCB & 70 & $0.034$ & $[0.026,\, 0.043]$ \\
Raster+Analyst & 70 & $0.287$ & $[0.257,\, 0.316]$ \\
GP-UCB+Analyst & 70 & $0.283$ & $[0.253,\, 0.311]$ \\
STEMAgent & 70 & $0.278$ & $[0.248,\, 0.305]$ \\
\botrule
\end{tabular}
\end{table}

\begin{table}[t]
\centering
\caption{Paired Wilcoxon signed-rank tests on matched (world, seed) episodes (defect census).}
\label{tab:wilcoxon}
\begin{tabular}{lrccc}
\toprule
\textbf{Comparison} & \textbf{$N$} & \textbf{$\Delta$ DEC-AUC} & \textbf{$p$-value} & \textbf{Sig.} \\
\midrule
Raster vs Raster+Analyst & 70 & $+0.235$ & $< 0.001$ & *** \\
Raster+Analyst vs GP-UCB+Analyst & 70 & $-0.004$ & $0.549$ &  \\
Raster+Analyst vs STEMAgent & 70 & $-0.009$ & $0.375$ &  \\
STEMAgent vs No Uncertainty & 30 & $+0.001$ & $0.903$ &  \\
STEMAgent vs No Planner & 30 & $+0.000$ & $0.855$ &  \\
\botrule
\end{tabular}
\end{table}

\paragraph{Equivalence testing (TOST).}
To formally claim that the three Analyst-equipped agents are equivalent rather than merely non-distinguishable, we run two one-sided Wilcoxon signed-rank tests (TOST) with equivalence margin $\Delta = 0.05$ DEC-AUC on the matched-seed N=70 paired arrays. A pair is declared equivalent if both one-sided $p$-values are below $0.05$, ruling out true mean differences of magnitude $\geq \Delta$. All three pairwise comparisons (Raster+Analyst vs.\ GP-UCB+Analyst, Raster+Analyst vs.\ STEMAgent, GP-UCB+Analyst vs.\ STEMAgent) pass equivalence at this margin.

\begin{table}[h]
\centering
\caption{Two one-sided Wilcoxon signed-rank tests (TOST) for equivalence among Analyst-equipped agents on defect census, with equivalence margin $\Delta = 0.05$ DEC-AUC. Pair counts are episodes matched by (world, seed). A pair is declared equivalent if both one-sided $p$-values are $< 0.05$.}
\label{tab:tost}
\footnotesize
\setlength{\tabcolsep}{4pt}
\begin{tabular}{llcccccc}
\toprule
\textbf{Agent A} & \textbf{Agent B} & $N$ & mean$(A{-}B)$ & std$(A{-}B)$ & $p_{\text{low}}$ & $p_{\text{up}}$ & \textbf{Equiv.?} \\
\midrule
Raster+Analyst & GP-UCB+Analyst & $70$ & $+0.0041$ & $0.0720$ & $0.0000$ & $0.0000$ & Yes \\
Raster+Analyst & STEMAgent & $70$ & $+0.0094$ & $0.0863$ & $0.0000$ & $0.0001$ & Yes \\
GP-UCB+Analyst & STEMAgent & $70$ & $+0.0053$ & $0.0774$ & $0.0000$ & $0.0000$ & Yes \\
\bottomrule
\end{tabular}
\end{table}

\section{PF GP-BO: Per-World Breakdown}
\label{app:kalinin}

The PF GP-BO baseline re-implements the autonomous-STEM active-learning workflow~\citep{ziatdinov2022hypothesis,kalinin2023ml4stem} using a BoTorch \texttt{SingleTaskGP}. The published pipeline uses learned tile embeddings or domain-specific descriptors; we substitute hand-crafted physics features (intensity variance, atom density from analyst peak detection, atom-intensity dispersion) for transparency and reproducibility. Acquisition is UCB ($\beta = 2.0$); the GP fit uses the same scaffolding as the GP-UCB baseline (\S\ref{sec:agents}). Evaluation covers all five easy worlds at \SI{5000}{\elementarycharge\per\angstrom\squared} with 10 seeds per (agent, task, world) combination (300 episodes total). The substituted feature set is the most likely source of the ${\sim}14\times$ perovskite-vs-non-perovskite gap; results should be read as ``GP-BO with hand-crafted features'' rather than as a faithful reproduction of the published learned-embedding variant.

\begin{table}[h]
\centering
\caption{PF GP-BO per-world DEC-AUC (mean over 10 seeds). The equipped variant pairs PF GP-BO navigation with the same trained Analyst ensemble used by other equipped agents in Table~\ref{tab:main}.}
\label{tab:kalinin_breakdown}
\small
\begin{tabular}{l@{\hspace{6pt}}c@{\hspace{6pt}}c@{\hspace{6pt}}c@{\hspace{6pt}}c@{\hspace{6pt}}c}
\toprule
\textbf{Task / Variant} & \ce{SrTiO3} & \ce{BaTiO3} & SiGe & GaN & Pt \\
\midrule
Defect census, raw       & $0.017$ & $0.042$ & $0.001$ & $0.005$ & $0.000$ \\
Defect census, equipped  & $0.377$ & $0.360$ & $0.043$ & $0.036$ & $0.000$ \\
Phase mapping, raw       & $0.908$ & $0.000$ & $0.000$ & $0.000$ & $0.908$ \\
Phase mapping, equipped  & $0.151$ & $0.157$ & $0.196$ & $0.137$ & $0.000$ \\
Targeted, raw            & $0.000$ & $0.000$ & $0.000$ & $0.000$ & $0.000$ \\
Targeted, equipped       & $0.808$ & $0.823$ & $0.054$ & $0.054$ & $0.000$ \\
\bottomrule
\end{tabular}
\end{table}

The equipped variant's defect-census and targeted scores show a $\sim$14$\times$ perovskite-vs-non-perovskite gap that mirrors what the in-house \texttt{Raster+Analyst} baseline shows (\S\ref{sec:generalisation}); the raw phase-mapping scores of $0.908$ on \ce{SrTiO3} and Pt reflect trivially-correct phase-0 predictions on single-phase worlds (no analyst means no phase prediction; default class 0 matches ground truth on worlds with no phase boundaries).

\section{Full Agent--Task Results}
\label{app:full_results}

Table~\ref{tab:main} in the main text reports the primary results.
Table~\ref{tab:agent_count} enumerates the 33 agent configurations by family; Table~\ref{tab:perworld} below gives per-world breakdowns.

\begin{table}[h]
\centering
\caption{Enumeration of the 33 agent configurations contributing to the reported results, grouped by family. ``Equipped'' agents pair the listed navigation policy with the trained Analyst ensemble (\S\ref{sec:agents}). VLM agents pair Raster or GP-UCB navigation with a vision--language analyst; the frontier-flagship sanity check (Claude Opus 4.7) is run with Raster only on defect census (SI~\ref{app:vlm_details}).}
\label{tab:agent_count}
\footnotesize
\begin{tabular}{p{0.27\linewidth}rp{0.55\linewidth}}
\toprule
\textbf{Family} & $N$ & \textbf{Members} \\
\midrule
Na\"{\i}ve baselines             & $3$ & Random, Raster, GP-UCB \\
Equipped baselines               & $2$ & Raster+Analyst, GP-UCB+Analyst \\
STEMAgent and ablations          & $5$ & STEMAgent, No Planner, No Uncertainty, Rule Planner, LLM Planner \\
PF GP-BO                 & $2$ & raw, +Analyst (SI~\ref{app:kalinin}) \\
Compressed sensing               & $6$ & CS-10/20/30\% $\times$ \{raw, +Analyst\} \\
Reinforcement learning           & $6$ & DQN, PPO, SAC $\times$ \{raw, +Analyst\} \\
Production-tier VLM analysts     & $8$ & \{Claude Haiku 4.5, GPT-5 mini, Llama 4 Scout, Gemini 2.0 Flash\} $\times$ \{Raster, GP-UCB\} \\
Frontier-flagship VLM sanity check & $1$ & Claude Opus 4.7 $\times$ Raster (SI~\ref{app:vlm_details}) \\
\midrule
\textbf{Total}                   & \textbf{33} & \\
\bottomrule
\end{tabular}
\end{table}

\begin{table}[h]
\centering
\caption{Per-world DEC-AUC on defect census. Perception gap is large and consistent; navigation is second-order.}
\label{tab:perworld}
\footnotesize
\setlength{\tabcolsep}{4pt}
\begin{tabular}{llccccccc}
\toprule
& & \multicolumn{3}{c}{\textbf{Heuristic Analyst}} & \multicolumn{4}{c}{\textbf{Trained Analyst}} \\
\cmidrule(lr){3-5} \cmidrule(lr){6-9}
\textbf{Material} & \textbf{Diff.} & \textbf{Random} & \textbf{Raster} & \textbf{GP-UCB} & \textbf{Rast.+A} & \textbf{GP+A} & \textbf{STEM} & \textbf{DQN+A} \\
\midrule
\ce{SrTiO3} & Easy   & $0.016$ & $0.044$ & $0.017$ & $0.455$ & $0.380$ & $0.361$ & $0.149$ \\
\ce{SrTiO3} & Med.   & $0.030$ & $0.109$ & $0.031$ & $0.306$ & $0.372$ & $0.367$ & $0.096$ \\
\ce{BaTiO3} & Easy   & $0.031$ & $0.048$ & $0.028$ & $0.355$ & $0.348$ & $0.318$ & $0.137$ \\
\ce{BaTiO3} & Med.   & $0.042$ & $0.059$ & $0.073$ & $0.359$ & $0.325$ & $0.349$ & $0.127$ \\
\ce{BaTiO3} & Hard   & $0.071$ & $0.097$ & $0.080$ & $0.340$ & $0.324$ & $0.293$ & $0.149$ \\
SiGe        & Easy   & $0.002$ & $0.003$ & $0.002$ & $0.070$ & $0.092$ & $0.087$ & $0.023$ \\
SiGe        & Med.   & $0.005$ & $0.005$ & $0.005$ & $0.126$ & $0.140$ & $0.171$ & $0.168$ \\
\bottomrule
\end{tabular}
\end{table}

\paragraph{Budget-sensitivity sweep.}
Table~\ref{tab:budget_sensitivity} reports the high-budget extension of the budget-sensitivity analysis discussed in \S\ref{sec:generalisation} (three crystalline easy worlds, 10 seeds per cell). The within-equipped ranking is preserved across the $50\times$ sweep and Raster+Analyst's margin grows with budget. SiGe stays at $0.05$--$0.08$ at every budget, exhibiting the cleanest material-specific perception ceiling.

\begin{table}[h]
\centering\small
\caption{DEC-AUC on defect census across dose budgets (mean$\pm$std, 10 seeds, three crystalline easy worlds). Coverage approximates the fraction of the 40$\times$40 tile grid acquired at the corresponding budget.}
\label{tab:budget_sensitivity}
\begin{tabular}{lcccc}
\toprule
\textbf{Agent} & \SI{1000}{} & \SI{5000}{} (baseline) & \SI{20000}{} & \SI{50000}{} \\
& ($\sim$0.6\%) & ($\sim$3\%) & ($\sim$12\%) & ($\sim$31\%) \\
\midrule
Raster (na\"{\i}ve) & $<0.04$         & $0.052$           & $<0.083$           & $<0.083$           \\
Raster+Analyst      & $0.158$--$0.162$ & $0.287\pm0.129$ & $0.340\pm0.164$    & $0.361\pm0.186$    \\
GP-UCB+Analyst      & $0.158$--$0.162$ & $0.283\pm0.122$ & $0.313\pm0.190$    & $0.321\pm0.192$    \\
STEMAgent           & $0.158$--$0.162$ & $0.278\pm0.120$ & $0.286\pm0.177$    & $0.314\pm0.183$    \\
\bottomrule
\end{tabular}
\end{table}

\section{Difficulty Scaling}
\label{sec:si:difficulty}

\begin{table}[h]
\centering\small
\caption{Physical parameters across \ce{BaTiO3} difficulty levels.}
\label{tab:bto_difficulty}
\begin{tabular}{lccc}
\toprule
Parameter & Easy & Medium & Hard \\
\midrule
O-vacancy rate           & $4.0\%$ & $3.0\%$ & $1.5\%$ \\
\# phase boundaries      & $1$ & $2$ & $3$ \\
Boundary sharpness $\sigma$ (px) & $80$ & $50$ & $25$ \\
Boundary waviness (frac.)& $0.00$ & $0.05$ & $0.10$ \\
Cluster fraction         & $0.60$ & $0.70$ & $0.80$ \\
Phonon configurations    & $4$ & $8$ & $16$ \\
\bottomrule
\end{tabular}
\end{table}

The \ce{BaTiO3} O-vacancy rate of $4\%/3\%/1.5\%$ for easy/medium/hard differs from the canonical $5\%/3\%/1\%$ in Table~\ref{tab:difficulty} of the main text because the phase-boundary structure already injects compositional variability that compounds with vacancies; equalising the resulting analyst-difficulty distributions across materials motivated the lower vacancy rates here.

\begin{table}[h]
\centering\small
\caption{DEC-AUC on \ce{BaTiO3} (mean$\pm$std, 10 seeds, \SI{5000}
{\elementarycharge\per\angstrom\squared} budget).}
\label{tab:difficulty_full}
\begin{tabular}{llccc}
\toprule
Task & Agent & Easy & Medium & Hard \\
\midrule
\multirow{4}{*}{Defect census}
 & Raster (na\"{\i}ve)   & $0.048\pm0.002$ & $0.059\pm0.005$ & $0.097\pm0.005$ \\
 & Raster+Analyst         & $0.355\pm0.002$ & $0.359\pm0.002$ & $0.340\pm0.004$ \\
 & GP-UCB+Analyst         & $0.348\pm0.055$ & $0.325\pm0.042$ & $0.324\pm0.079$ \\
 & STEMAgent              & $0.318\pm0.076$ & $0.349\pm0.041$ & $0.293\pm0.095$ \\
\midrule
\multirow{4}{*}{Phase mapping}
 & Raster (na\"{\i}ve)   & $0.000$ & $0.000$ & $0.000$ \\
 & Raster+Analyst         & $0.149\pm0.004$ & $0.197\pm0.005$ & $0.159\pm0.016$ \\
 & GP-UCB+Analyst         & $0.149\pm0.051$ & $0.234\pm0.031$ & $0.168\pm0.012$ \\
 & STEMAgent              & $0.168\pm0.022$ & $0.221\pm0.025$ & $0.167\pm0.015$ \\
\bottomrule
\end{tabular}
\end{table}

\paragraph{Specimen protocol.}
The three \ce{BaTiO3} interface worlds share substrate geometry
(BaTiO$_3$\,[001] with mixed cubic/tetragonal domains) and differ only in
parameters that control defect detectability and structural complexity
(Table~\ref{tab:bto_difficulty}). O-vacancy density decreases from $4.0\%$
to $1.5\%$, the number of cubic/tetragonal phase boundaries rises from one
to three, the Gaussian boundary sharpness $\sigma$ drops from 80 to
25\,px, and boundary waviness grows from $0$ to $10\%$ of scan width. The
HAADF simulation ramps thermal-diffuse-scattering realism in parallel,
from 4 to 16 frozen-phonon configurations. All other parameters are held
fixed across difficulties: \SI{200}{\kilo\electronvolt} beam energy,
\SI{21}{\milli\radian} convergence semi-angle, a $68$--$\SI{200}
{\milli\radian}$ HAADF detector, \SI{128}\,px tiles, a \SI{5000}
{\elementarycharge\per\angstrom\squared} budget, and a 200-step cap.

\paragraph{Per-agent, per-task results.}
Table~\ref{tab:difficulty_full} reports DEC-AUC mean$\pm$std across ten
seeds per (agent, world) pair. Two aggregate patterns emerge. On defect
census, equipped agents span only $0.293$--$0.359$ (a $22\%$ relative
range) while na\"{\i}ve baselines stay below $0.10$. On phase mapping,
na\"{\i}ve agents return exactly zero at every difficulty (none of
random, raster, or GP-UCB exploration produces a phase map without the
trained analyst), while all three equipped agents peak at the medium
world ($0.197$--$0.234$) and decline at both endpoints.

\paragraph{Non-monotonic phase-mapping response.}
The medium-peak pattern appears consistently in all three equipped
agents, indicating it is driven by the shared phase identifier rather
than by the navigation policy. Our phase-mapping metric is macro-averaged
IoU across the two phase labels. We offer a plausible decomposition: in
the easy world a single boundary makes the IoU dominated by one binary
decision, so mislabeling either side degrades IoU sharply; the medium
world introduces a second boundary, increasing within-scan contrast
diversity and raising the plateau; the hard world re-introduces
degradation as three wavy boundaries (with sharper $\sigma$ and
lower-SNR imaging conditions) outstrip what the phase identifier can
resolve under the \SI{5000}{\elementarycharge\per\angstrom\squared}
budget. A controlled ablation decoupling boundary count from imaging SNR
would be needed to attribute the effect cleanly.

\paragraph{Narrowing of the defect-census gap.}
Na\"{\i}ve Raster's defect-census AUC rises from $0.048$ (easy) to
$0.097$ (hard), and GP-UCB's rises from $0.028$ to $0.080$, while their
coverage policies are unchanged. The defect-census score is
$\max(\text{count-F1}, \text{spatial-F1})$, and the na\"{\i}ve analyst
uses a fixed $\pm 1.5$ z-score threshold on tile peak intensities, so
its reported defect count is approximately independent of ground-truth
vacancy density. Sparser hard-world ground truth therefore brings the
reported count incidentally closer to truth on the count-F1 branch. The
perception gap on defect census consequently narrows from $7.4\times$
(easy) to $3.5\times$ (hard), while the gap on phase mapping remains
undefined (na\"{\i}ve $\equiv 0$). Equipped agents' defect-census scores
change by at most $0.056$ across the sweep (STEMAgent easy-to-hard), so
the narrowing is driven almost entirely by the rising na\"{\i}ve floor,
not by degrading perception.

\paragraph{Intra-agent stability.}
At each difficulty, Raster+Analyst has the lowest seed-to-seed variance
($\sigma \leq 0.004$ on defect census), followed by GP-UCB+Analyst
($\sigma \leq 0.079$), with STEMAgent highest ($\sigma$ up to $0.095$).
The ordering is stable across all three difficulty levels and reflects a
coverage-versus-adaptation tradeoff: under a tight
\SI{5000}{\elementarycharge\per\angstrom\squared} budget, deterministic
sweeps inherit only the analyst's aggregate variance, whereas adaptive
agents additionally inherit variance from their own exploration
policies.

\section{Computational Resources}
\label{app:compute}

\paragraph{World Generation.}
Easy and medium worlds required ${\sim}$2--6 hours on a single NVIDIA A100 (80~GB).
Hard worlds require $> \SI{100}{\giga\byte}$ GPU memory and were generated on H200 GPUs (141~GB); the largest worlds used four parallel H200 instances, requiring ${\sim}$14--17 hours per world.

\paragraph{Experiments.}
The main suite ran on two RTX 4090 instances (seed-sharded); VLM/LLM comparisons ran on an H200; expansion experiments ran on a second A100. The full episode budget is bounded by the configuration files released with the benchmark and the cross-material transfer programme (3 agents $\times$ 5 train materials $\times$ 5 eval materials $\times$ 5 seeds $=$ 375 episodes), summing to approximately 8{,}000 episodes. Per-experiment breakdown:

\begin{itemize}\setlength{\itemsep}{0pt}
\item \textbf{Main suite} ($2{,}820$): defect census ($1{,}120$), phase mapping ($480$), targeted ($320$), particle census ($450$), trained ablation ($450$).
\item \textbf{VLM \& LLM} ($1{,}875$): original VLM sweep ($900$; GPT-5 mini chronic parse failures, see SI~\ref{app:vlm_diagnostics}), retry-harness rerun ($760$ effective episodes after exclusion), Claude Opus 4.7 frontier-flagship sanity check ($15$; defect census only, see SI~\ref{app:vlm_details}), LLM planner ($200$).
\item \textbf{PF GP-BO and compressed sensing} ($660$): PF GP-BO ($300$), CS-10/20/30\% raw $+$ equipped ($360$).
\item \textbf{Budget and Pareto sweeps} ($360$): 1k/3k/10k Pareto ($180$), 20k/50k budget ($180$).
\item \textbf{Cross-material transfer} ($375$).
\item \textbf{Replay validation and trajectory capture} ($48$).
\item \textbf{Real-data sanity check} ($30$): five Sm-doped \ce{BiFeO3} compositions (Sm $\in\{0, 7, 10, 13, 20\}\%$)~\citep{ghosh2021bfo}, surfaced via the 2024 Mic-hackathon team~14 notebook~\citep{pratiush2025michackathon}, 2 agents $\times$ 5 worlds $\times$ 3 seeds on phase mapping (SI~\ref{app:realdata}).
\item \textbf{Targeted ablations} ($1{,}822$): defect-census 5-seed sweep ($350$), zoom/dwell ablation ($225$), acquisition-function comparison ($150$), composite task ($75$), adaptive transfer ($60$), declare ablation ($60$), random/single-BO equipped runs ($500$), trained difficulty scaling ($120$), trained targeted ($200$), RL baseline ($50$), preflight ($32$).
\end{itemize}

\noindent The bullets sum to $7{,}990$ (rounded to ${\sim}8{,}000$ in the headline figure). Every count is bounded by a released configuration file or by the cross-material transfer programme.

\noindent No experiment required more than 80~GB; agent inference peaks at ${\sim}$40~GB with 4 parallel workers.

\paragraph{Per-agent wall-clock latency.}
Table~\ref{tab:latency} reports per-episode wall-clock time and median per-acquired-tile latency aggregated from \texttt{elapsed\_seconds} across all readable result files. Per-tile latency on the trained-analyst agents is dominated by U-Net inference and BoTorch GP fitting; PPO is an outlier because it spends additional wall-clock on per-step policy updates. VLM rows are dominated by OpenRouter call latency at \texttt{max\_tokens}=1024; the GPT-5 mini rows additionally absorb 3-retry overhead per failed call (see SI~\ref{app:vlm_diagnostics}), inflating per-tile latency to $45$--$90$~s. Pareto-sweep wall-clock data are not yet included pending recovery of those JSONs.

\begin{table}[h]
\centering
\caption{Per-agent wall-clock latency aggregated from \texttt{elapsed\_seconds} across all readable result files. Per-tile median is the median of $\mathrm{elapsed}/\mathrm{tiles\_acquired}$ over episodes that acquired at least one tile. VLM rows reflect end-to-end OpenRouter call latency at \texttt{max\_tokens}=1024 with retry overhead; the Claude Opus 4.7 row aggregates the frontier-flagship sanity check (defect census only). Pareto-sweep JSONs are unavailable locally.}
\label{tab:latency}
\footnotesize
\setlength{\tabcolsep}{4pt}
\begin{tabular}{lrrrr}
\toprule
\textbf{Agent} & $N$ episodes & Mean episode (s) & Median episode (s) & Median per-tile (ms) \\
\midrule
Random & $156$ & $123.6$ & $81.2$ & $1691.3$ \\
Raster & $231$ & $614.7$ & $159.0$ & $3179.6$ \\
GP-UCB & $231$ & $518.0$ & $166.8$ & $3336.2$ \\
Raster+Analyst & $265$ & $681.5$ & $205.7$ & $3505.4$ \\
GP-UCB+Analyst & $255$ & $561.8$ & $250.9$ & $3420.9$ \\
STEMAgent & $276$ & $200.1$ & $116.2$ & $2222.8$ \\
STEMAgent (No Planner) & $240$ & $137.4$ & $116.1$ & $2369.9$ \\
STEMAgent (No Uncertainty) & $240$ & $142.0$ & $122.6$ & $2554.6$ \\
STEMAgent (Rule Planner) & $240$ & $146.2$ & $127.7$ & $2659.7$ \\
STEMAgent (LLM Planner) & $210$ & $169.2$ & $138.4$ & $2883.7$ \\
PF GP-BO & $150$ & $38.6$ & $38.2$ & $763.9$ \\
PF GP-BO+Analyst & $150$ & $42.7$ & $40.2$ & $804.3$ \\
DQN & $150$ & $164.7$ & $131.7$ & $14631.1$ \\
DQN+Analyst & $150$ & $167.6$ & $137.0$ & $15412.1$ \\
PPO & $150$ & $169.6$ & $139.9$ & $69943.5$ \\
PPO+Analyst & $150$ & $171.8$ & $143.5$ & $71742.0$ \\
SAC & $150$ & $174.0$ & $146.5$ & $3021.6$ \\
SAC+Analyst & $150$ & $178.4$ & $153.7$ & $3202.4$ \\
Raster+Claude Haiku 4.5 & $75$ & $1715.9$ & $1128.7$ & $22574.6$ \\
Raster+Gemini 2.0 Flash & $74$ & $1353.1$ & $1172.2$ & $23443.8$ \\
Raster+Llama 4 Scout & $64$ & $1498.1$ & $799.9$ & $15998.7$ \\
Raster+GPT-5 mini & $15$ & $2770.8$ & $2269.9$ & $45398.5$ \\
Raster+Claude Opus 4.7 & $15$ & $353.1$ & $365.9$ & $7318.3$ \\
GP-UCB+Claude Haiku 4.5 & $75$ & $1504.2$ & $1053.1$ & $21062.2$ \\
GP-UCB+Gemini 2.0 Flash & $75$ & $1789.1$ & $1246.1$ & $24921.8$ \\
GP-UCB+Llama 4 Scout & $67$ & $1932.1$ & $1073.0$ & $21459.3$ \\
GP-UCB+GPT-5 mini & $15$ & $4456.4$ & $4506.4$ & $90128.7$ \\
\bottomrule
\end{tabular}
\end{table}

\section{Exploration Patterns}
\label{app:exploration}

\begin{figure}[h]
\centering
\begin{tikzpicture}
\node[inner sep=4pt, fill=gray!6, rounded corners=3pt, draw=gray!25, line width=0.4pt] {%
\includegraphics[width=0.95\textwidth]{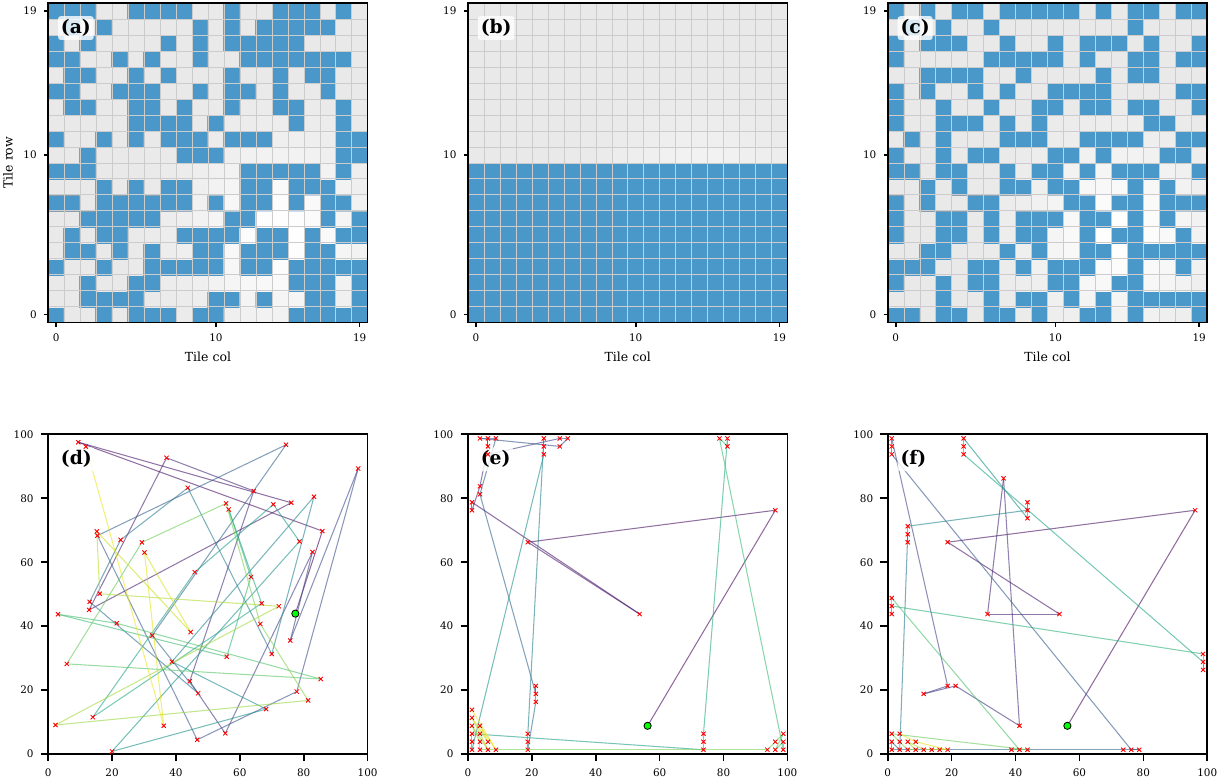}%
};
\end{tikzpicture}
\caption{Spatial exploration and acquisition trajectories.
\textbf{(a--c)}~Coverage maps at 50\% dose on Si/Ge easy for Random, Raster, and GP-UCB.
\textbf{(d--f)}~Acquisition trajectories on \ce{SrTiO3} easy for Random+CNN, GP-UCB+CNN, and STEMAgent.}
\label{fig:appendix_analyses}
\end{figure}

\paragraph{Exploration Patterns.}
Random produces uniform coverage, Raster scans row-by-row, GP-UCB concentrates on high-variance regions.
Despite distinct strategies, all three na\"{\i}ve baselines achieve near-zero DEC-AUC, confirming that navigation without trained perception contributes negligibly.
Equipped agents produce visually distinct trajectories but achieve similar scores (Random+CNN $0.301$, STEMAgent $0.377$, GP-UCB+CNN $0.394$ on \ce{SrTiO3}), directly supporting the ``navigation is noise'' finding.

\section{Ablation and Planner Comparison}
\label{app:ablation}

\begin{table}[h]
\centering
\caption{Ablation on medium-difficulty worlds (3 worlds $\times$ 10 seeds). DEC-AUC mean $\pm$ std.}
\label{tab:ablation}
\begin{tabular}{l@{\hspace{8pt}}c@{\hspace{8pt}}c@{\hspace{8pt}}c}
\toprule
\textbf{Agent Variant} & \textbf{Defect Census} & \textbf{Phase Mapping} & \textbf{Targeted} \\
\midrule
STEMAgent (full)      & $0.283 \pm 0.122$ & $0.374 \pm 0.363$ & $0.654 \pm 0.341$ \\
\quad w/o Planner     & $0.283 \pm 0.116$ & $0.402 \pm 0.367$ & $0.636 \pm 0.348$ \\
\quad w/o Uncertainty & $0.284 \pm 0.106$ & $0.404 \pm 0.366$ & $0.652 \pm 0.332$ \\
\quad Rule Planner    & $0.304 \pm 0.106$ & $0.400 \pm 0.370$ & $0.634 \pm 0.361$ \\
\quad LLM Planner     & $0.292 \pm 0.111$ & $0.410 \pm 0.363$ & $0.667 \pm 0.333$ \\
\bottomrule
\end{tabular}
\end{table}

No single component is clearly consequential: all variants score within $0.021$ on defect census ($0.283$--$0.304$) and within $0.036$ on phase mapping ($0.374$--$0.410$).
Planner removal has zero impact on defect census ($0.283 \to 0.283$).
High variance ($\sigma/\mu > 0.4$) means no variant is statistically distinguishable.

\begin{table}[h]
\centering
\caption{Planner variant comparison across five easy worlds (10 seeds). All variants are indistinguishable.}
\label{tab:llmplanner}
\begin{tabular}{lcc}
\toprule
\textbf{Planner Variant} & \textbf{Defect Census} & \textbf{Phase Mapping} \\
\midrule
FSM (full)    & $0.255$ & $0.168$ \\
LLM Planner   & $0.263$ & $0.168$ \\
Rule Planner  & $0.273$ & $0.161$ \\
No Planner    & $0.270$ & $0.165$ \\
\bottomrule
\end{tabular}
\end{table}

\section{Compressed Sensing}
\label{app:compressed}

Compressed sensing~\citep{candes2006robust} is a complementary paradigm to active acquisition: rather than choosing \emph{where} to measure, it acquires a sparse random subset and reconstructs the rest via inpainting. We implemented this as three \texttt{CompressedSensingAgent} variants with random-uniform masks at 10\%, 20\%, and 30\% nominal coverage. During the episode the agent acquires only unmasked tiles. After the episode terminates, the unacquired tiles are reconstructed via total-variation inpainting (Chambolle's algorithm~\citep{chambolle2004algorithm}); the trained analyst is then run over the reconstructed tiles using a separate analyst instance to keep real-tile and reconstructed-tile intensity statistics from contaminating the median-split phase-bimodality test.

\paragraph{Per-coverage and per-world results.}
Table~\ref{tab:cs_breakdown} reports per-world DEC-AUC for the three equipped variants on defect census across three crystalline easy worlds (10 seeds each). Coverage fraction has minimal effect on aggregate score because the dose budget caps acquisitions at $\sim$50 tiles ($\sim$3\% of the 40$\times$40 grid), so the 10/20/30\% masks are equally budget-limited and the variable being tested is the spatial distribution of acquired tiles. Per-world, the perovskites (\ce{SrTiO3}, \ce{BaTiO3}) tolerate inpainting at $0.18$--$0.21$ DEC-AUC, while SiGe drops to $0.06$--$0.10$ because Si and Ge atomic columns differ in HAADF intensity, and TV smoothing erases that signal.

\begin{table}[ht]
\centering
\caption{Compressed-sensing equipped variants, defect census DEC-AUC (mean $\pm$ std, 10 seeds per cell). The dose budget caps acquisitions at $\sim$50 tiles for all three variants, so the difference between rows is the spatial distribution of which tiles are masked, not how many are acquired.}
\label{tab:cs_breakdown}
\begin{tabular}{lccc}
\toprule
\textbf{Coverage} & \ce{SrTiO3} & \ce{BaTiO3} & SiGe \\
\midrule
10\% (CS-10\%+Analyst) & $0.177 \pm 0.002$ & $0.205 \pm 0.005$ & $0.064 \pm 0.001$ \\
20\% (CS-20\%+Analyst) & $0.210 \pm 0.002$ & $0.212 \pm 0.003$ & $0.102 \pm 0.032$ \\
30\% (CS-30\%+Analyst) & $0.187 \pm 0.001$ & $0.213 \pm 0.010$ & $0.076 \pm 0.001$ \\
\bottomrule
\end{tabular}
\end{table}

\paragraph{Domain-shift caveat.}
The U-Net atom finder was trained on raw HAADF tiles. TV inpainting is a denoising operation that smooths high-frequency content by construction, including the atomic-column structure that the U-Net was trained to localise. The CS+Analyst rows therefore conflate two effects: budget-limited measurement and a domain shift in the analyst's input distribution. A fair upper bound on the CS paradigm would require either retraining the analyst on TV-inpainted tiles or substituting a reconstruction algorithm that preserves atomic-column structure (e.g., dictionary learning on raw HAADF patches); both are natural extensions of the present implementation.

\paragraph{Phase mapping.}
Phase-mapping numbers in Table~\ref{tab:main} have wide confidence intervals (std $0.18$--$0.26$) because the trivial single-phase worlds (\ce{SrTiO3}, SiGe with the chosen ground-truth labelling) score near $1.0$ when the analyst defaults to ``predict phase 0,'' while \ce{BaTiO3} with its real phase boundary scores near $0$. Aggregate IoU therefore reflects the worlds-included composition rather than reconstruction quality and should not be over-interpreted.

\section{VLM Implementation Details}
\label{app:vlm_details}

\begin{table}[h]
\centering
\caption{VLM API configuration.}
\label{tab:vlmconfig}
\resizebox{\textwidth}{!}{%
\begin{tabular}{ll}
\toprule
\textbf{Parameter} & \textbf{Value} \\
\midrule
API provider       & OpenRouter (\texttt{openrouter.ai/api/v1}) \\
Models             & \texttt{anthropic/claude-haiku-4.5}, \texttt{openai/gpt-5-mini}, \\
                   & \texttt{meta-llama/llama-4-scout}, \texttt{google/gemini-2.0-flash-001}, \\
                   & \texttt{anthropic/claude-opus-4.7} (frontier-flagship sanity check) \\
max\_tokens        & 1024 \\
Temperature        & Provider default (not explicitly set) \\
image\_detail      & \texttt{high} \\
Image preprocessing & $2\times$ nearest-neighbour upscale ($128 \to 256$~px), \\
                   & normalised to $[0, 255]$, PNG-encoded \\
Retry policy       & Up to 3 attempts wrapped around the OpenRouter call boundary; \\
                   & retries on \texttt{json.JSONDecodeError} only, with 1\,s sleep between attempts; \\
                   & non-retryable errors return empty findings \\
JSON robustness    & Strips markdown code fences (\texttt{```json} preferred, generic \texttt{```} fallback), \\
                   & extracts last balanced \texttt{\{...\}} substring; falls back to empty on parse error \\
Empty content guard & Empty or null response content raises a JSON parse error \\
                   & so the retry path catches it as a transient failure \\
\bottomrule
\end{tabular}}
\end{table}

\paragraph{System prompt} (verbatim):
\begin{quote}\small
\textit{You are an expert materials scientist analyzing HAADF-STEM (High-Angle Annular Dark-Field Scanning Transmission Electron Microscopy) images of crystalline materials.}

\textit{In HAADF-STEM images: Bright spots are atomic columns (intensity $\sim Z^{1.7}$, heavier atoms appear brighter). Vacancies appear as MISSING or abnormally DIM spots in the lattice. Substitutional defects appear as spots that are significantly BRIGHTER or DIMMER than their neighbors (different atomic number). The image is a $128\times128$ pixel tile from a larger specimen.}

\textit{Analyze the tile and respond with ONLY a JSON object (no markdown, no explanation).}
\end{quote}

\paragraph{User prompt} (verbatim):
\begin{quote}\small
\textit{Analyze this HAADF-STEM tile image. Return a JSON object with:}
\texttt{\{"n\_atoms": <int>, "defects": [\{"row": <0--127>, "col": <0--127>, "type": "vacancy"|"substitution"\}], "phase": <int>, "mean\_intensity": <float 0--1>, "confidence": <float 0--1>\}}

\textit{If no defects are visible, return an empty defects array. Focus on clear, unambiguous defects only---do not over-report.}
\end{quote}

\paragraph{LLM Planner.}
The LLM Planner uses Gemini 3 Flash Preview (\texttt{google/gemini-3-flash-preview}) via OpenRouter with temperature $0.3$ and a LangGraph-based tool-calling agent (recursion limit $= 7$).
The planner decides FSM mode transitions (SURVEY, INVESTIGATE, CHARACTERIZE, CENSUS, TERMINATE) based on the current experiment state.
When the recursion limit is reached, the planner falls back to rule-based mode selection.

\paragraph{Frontier-flagship sanity check (Claude Opus 4.7).}
To address whether the production-tier perception gap reflects a model-tier limitation rather than VLM perception in general, we ran a single frontier-flagship configuration on the headline task: \texttt{anthropic/claude-opus-4.7} via OpenRouter with the same prompts, image preprocessing, retry harness, and \texttt{max\_tokens}=1024 cap as the production-tier sweep, paired with raster navigation on defect census across the five easy worlds with three seeds each (15 episodes total, $750$ underlying tile calls). Opus 4.7 returned cleanly fenced JSON on every call (zero parse failures, zero retries triggered). Per-world DEC-AUC was $0.019\pm0.002$ (\ce{SrTiO3}), $0.037\pm0.002$ (\ce{BaTiO3}), $0.003\pm0.000$ (SiGe), $0.005\pm0.001$ (GaN), and $0.908\pm0.000$ (Pt nanoparticles), giving a four-world crystalline mean of $0.016$---inside the $0.006$--$0.019$ range observed for the production-tier set. The Pt result of $0.908$ matches the trivially-correct na\"{\i}ve raster baseline, consistent with the ``default no-anomaly'' explanation in \S\ref{sec:tasks} (${\sim}85\%$ of Pt tiles contain no particle, so an empty defects array scores high $F_{1,\text{detect}}$ regardless of perception). The result is consistent with the production-tier sweep: the perception gap relative to the trained CNN is not closed by switching to a frontier-tier model under API-default settings.

\section{VLM Parse-Failure Diagnostics}
\label{app:vlm_diagnostics}

The CoT-with-retry implementation logs each attempt outcome (model, attempt index, success flag, parse-failure reason, response excerpt) in a per-instance diagnostics list cleared on every \texttt{reset()}. The aggregate failure counts across the full run are summarised in Table~\ref{tab:vlm_parse_diag}.

\begin{table}[h]
\centering
\caption{Retry-exhausted parse failures per VLM across all evaluated combos. ``Failures'' counts only events where the third and final retry attempt also failed to produce a valid JSON object; first/second-attempt failures that succeeded on retry are not counted. ``Episodes'' is the number of evaluation episodes the model contributed to the final analysis.}
\label{tab:vlm_parse_diag}
\resizebox{\textwidth}{!}{%
\begin{tabular}{lcccc}
\toprule
\textbf{Model} & \textbf{Episodes} & \textbf{Retry-exhausted failures} & \textbf{Failure rate} & \textbf{Status in main results} \\
\midrule
Claude Haiku 4.5    & 150 & $0$                & $0\%$        & Included \\
Gemini 2.0 Flash    & 149 & $3$                & ${\sim}0\%$  & Included \\
Llama 4 Scout       & 131 & $0$                & $0\%$        & Included \\
GPT-5 mini          & 30  & $123$              & ${\sim}100\%$ & Excluded (see below) \\
\bottomrule
\end{tabular}}
\end{table}

\paragraph{GPT-5 mini exclusion.}
GPT-5 mini consistently returned empty response content from OpenRouter under our $\texttt{max\_tokens}=1024$ cap. Inspection of non-empty responses showed two distinct issues. The first is genuine API-side truncation: the model spends its token budget on internal reasoning before emitting the structured output and is cut off mid-object, producing payloads such as \texttt{\{"n\_atoms":186,"defects":[],"phase":0,"mean\_intensity":0.11,"confidence":0.78} (no closing brace). The second was a parser-side artefact in our JSON-extraction helper: when a non-empty response wrapped a nested \texttt{defects:[\{...\}]} array, the previous last-brace fallback latched onto the inner defect object's brace and produced fragments like \texttt{\{"row": 64, "col": 64, "type": "vacancy"\}],"phase": 0,...} that resembled API truncation but were actually parsing errors on cleanly-emitted JSON. We patched the helper to use a brace-matching scanner that returns the outermost balanced object and re-ran a 30-combo gap-fill (2 navigators $\times$ 3 tasks $\times$ 5 worlds, 5 seeds each); the parse-failure rate remained near $100\%$, dominated by genuinely-empty content and genuinely-truncated payloads, so no GPT-5 mini episodes were recovered. The original API-side diagnosis therefore stands: under $\texttt{max\_tokens}=1024$, GPT-5 mini cannot reliably emit the structured payload on real material tiles, and the retry path catches this case (the empty-content guard in SI~\ref{app:vlm_details} converts an empty response into a JSON parse error) but every retry produced the same outcome. After 30 episodes of the original sweep confirmed this was chronic API behaviour rather than transient noise we excluded GPT-5 mini from the headline analysis; the partial results are retained in the supplementary archive for completeness but do not contribute to Tables~\ref{tab:main}, \ref{tab:vlmdetail}, or \ref{tab:vlmtargeted}. Raising \texttt{max\_tokens} above $1024$ may resolve the truncation issue for reasoning-heavy models; we leave this configuration sweep to future work since the goal of the present comparison is API-default production-tier-model behaviour rather than per-model tuning.

\paragraph{Working-model failures.}
Gemini 2.0 Flash exhibited 3 retry-exhausted failures across all $149$ evaluation episodes ($\sim$7,500 underlying VLM tile calls at one call per acquired tile), a per-episode failure rate of ${\sim}0.02\%$. The 3 failures correlated with one OpenRouter \texttt{INVALID\_ARGUMENT} response from Google's API and two cases of malformed JSON output. Claude Haiku and Llama 4 Scout had zero retry-exhausted failures, indicating that the JSON-extraction logic (fenced block preferred, generic-fence fallback, last-balanced-braces fallback) is sufficient for these providers' typical output formats.

\section{VLM Per-World Breakdown}
\label{app:vlm}
Table~\ref{tab:vlmdetail} compares DEC-AUC of vision-language model (VLM) agents on defect census across the four crystalline easy worlds (\ce{SrTiO3}, \ce{BaTiO3}, SiGe, GaN) and the Pt nanoparticles easy world, under raster navigation. The results show a strong domain dependence: the trained CNN performs well on crystalline worlds ($0.247$) but fails on nanoparticles ($0.000$), whereas the working production-tier VLMs (Claude Haiku, Gemini 2.0, Llama 4) achieve high performance on nanoparticles (up to $0.908$) but near-zero performance on crystallography. Aggregated over all five worlds these opposing strengths result in similar overall DEC-AUC values for the CNN ($0.198$) and the strongest VLM ($0.186$ for Claude Haiku), with Raster+CNN slightly higher due to its stronger performance on the four crystalline worlds, while the heuristic raster baseline remains substantially lower. GPT-5 mini scored $0.000$ on every world due to chronic empty-content responses (SI~\ref{app:vlm_diagnostics}); we report it for completeness but exclude it from the headline comparison. Frontier-flagship Claude Opus 4.7 (15 episodes, no retries triggered) sits in the same crystalline range as the production tier ($0.016$ averaged over the four worlds vs.\ $0.006$--$0.019$ for the production-tier set) and matches the trivially-correct na\"{\i}ve baseline of $0.908$ on Pt nanoparticles, indicating the perception gap is not a tier-of-model artefact (SI~\ref{app:vlm_details}).

Table~\ref{tab:vlmtargeted} reports DEC-AUC on the targeted task across all five easy worlds. Here Raster+CNN clearly outperforms every VLM, achieving $0.392$ compared to near-zero performance for all VLMs. The targeted task demands spatially precise localisation rather than scene-level classification, so the inversion that benefits VLMs on Pt nanoparticles in defect census does not transfer.

\begin{table}[h]
\centering
\caption{VLM per-world DEC-AUC on defect census (raster navigation, 5 seeds, CoT-with-retry; SI~\ref{app:vlm_diagnostics}). Crystalline worlds = \ce{SrTiO3}, \ce{BaTiO3}, SiGe, GaN.}
\label{tab:vlmdetail}
\begin{tabular}{lccc}
\toprule
\textbf{Agent} & \textbf{Crystalline (4 worlds)} & \textbf{Pt Nanopart.} & \textbf{All Worlds} \\
\midrule
Raster+CNN          & $0.247$ & $0.000$ & $0.198$ \\
Raster+Claude Opus 4.7 & $0.016$ & $0.908$ & $0.194$ \\
Raster+Claude Haiku & $0.006$ & $0.908$ & $0.186$ \\
Raster+Gemini 2.0   & $0.019$ & $0.628$ & $0.141$ \\
Raster+Llama 4      & $0.019$ & $0.458$ & $0.041$ \\
Raster+GPT-5 mini   & $0.000$ & $0.000$ & $0.000$ \\
Raster (heuristic)  & $0.027$ & $0.000$ & $0.021$ \\
\bottomrule
\end{tabular}
\end{table}

\begin{table}[h]
\centering
\caption{VLM DEC-AUC on the targeted task (all 5 easy worlds, raster navigation, 5 seeds).}
\label{tab:vlmtargeted}
\begin{tabular}{lccc}
\toprule
\textbf{Agent} & \textbf{Crystalline (4 worlds)} & \textbf{Pt Nanopart.} & \textbf{All Worlds} \\
\midrule
Raster+CNN          & $0.490$ & $0.000$ & $0.392$ \\
Raster+Claude Haiku & $0.004$ & $0.000$ & $0.004$ \\
Raster+Gemini 2.0   & $0.019$ & $0.000$ & $0.016$ \\
Raster+Llama 4      & $0.020$ & $0.000$ & $0.018$ \\
Raster+GPT-5 mini   & $0.000$ & --- & $0.000$ \\
\bottomrule
\end{tabular}
\end{table}

\section{Reinforcement Learning Baseline Details}
\label{app:rl}

Three algorithms are trained via Stable-Baselines3 for 50{,}000 steps each across five easy worlds:

\textbf{DQN}~\citep{mnih2015human}: Discrete $8{\times}8$ spatial grid $\times$ 3 action types (192 actions).
Learning rate $10^{-4}$, replay buffer 50{,}000, batch size 64, exploration fraction 0.3, $\gamma{=}0.99$.

\textbf{PPO}~\citep{schulman2017proximal}: Same discrete action space.
Learning rate $3{\times}10^{-4}$, rollout length 2{,}048, 10 epochs per update, clip range 0.2, entropy coefficient 0.01, GAE $\lambda{=}0.95$.

\textbf{SAC}~\citep{haarnoja2018soft}: Continuous $\text{Box}(2)$ action space (normalised $x,y$ coordinates).
Learning rate $3{\times}10^{-4}$, replay buffer 50{,}000, batch size 64, soft-target Polyak coefficient $\tau_\text{pol}{=}0.005$ (renamed from $\tau$ to avoid collision with the DEC hitting time of \S\ref{sec:problem}), automatic entropy tuning.

\begin{table}[h]
\centering
\caption{RL per-world DEC-AUC on defect census (easy worlds, 10 seeds).}
\label{tab:dqnresults}
\begin{tabular}{lcccccc}
\toprule
\textbf{World} & \textbf{DQN} & \textbf{DQN+A} & \textbf{PPO} & \textbf{PPO+A} & \textbf{SAC} & \textbf{SAC+A} \\
\midrule
\ce{SrTiO3} easy & $0.064$ & $0.149$ & $0.003$ & $0.001$ & $0.022$ & $0.381$ \\
\ce{BaTiO3} easy & $0.172$ & $0.137$ & $0.002$ & $0.000$ & $0.026$ & $0.295$ \\
SiGe easy        & $0.005$ & $0.023$ & $0.001$ & $0.004$ & $0.002$ & $0.046$ \\
\midrule
\textbf{Mean}    & $\mathbf{0.080}$ & $\mathbf{0.103}$ & $\mathbf{0.002}$ & $\mathbf{0.002}$ & $\mathbf{0.017}$ & $\mathbf{0.241}$ \\
\bottomrule
\end{tabular}
\end{table}

\section{Datasheet for \stemgym{}}
\label{app:datasheet}

We follow the standard datasheets-for-datasets framework~\citep{gebru2021datasheets}.

\paragraph{Motivation.}
\stemgym{} is the first Gymnasium-compatible benchmark for evaluating dose-efficient autonomous scanning transmission electron microscopy agents.
No prior benchmark combines physics-based STEM simulation with ground-truth annotations and a standardised agent interface.
The dataset was created by the paper authors with no external dataset-specific funding.

\paragraph{Composition.}
The dataset comprises 15 benchmark worlds (five material systems at three difficulty levels) plus two ancillary HDF5 files used internally: a synthetic test world for unit testing and a synthetic replay world for cross-parameterisation consistency checks (Table~\ref{tab:datasheet_composition}). The abstract count of ``15 annotated HAADF-STEM worlds'' refers to the benchmark worlds only; the ancillary worlds are not used for any reported result.
Each world contains: (i)~a low-resolution overview image, (ii)~a grid of $128{\times}128$~pixel HAADF-STEM tiles with 4-pixel overlap (stride 124), (iii)~ground-truth annotations, and (iv)~a valid-region mask.
Ground truth includes per-atom position (nm), type (0\,=\,pristine, 1\,=\,vacancy, 2\,=\,substitution), defect mask, defect-type strings, and, where applicable, a pixel-wise phase map (\ce{BaTiO3}, GaN).
No confidential data or personally identifiable information is present; all content is computationally generated.
Tiles within a world are \emph{not} independent: they share a common crystal structure and exhibit spatial correlations.

The release also includes six model checkpoints (AtomFinderUNet ensemble, DefectClassifierCNN, PhaseIdentifierResNet, and DQN/PPO/SAC RL agents; ${\sim}$107~MB total) and 16 material-specific transfer checkpoints (${\sim}$420~MB).

\begin{table}[h]
\centering
\caption{Dataset composition per material system.}
\label{tab:datasheet_composition}
\begin{tabular}{llcccc}
\toprule
\textbf{Material} & \textbf{Zone Axis} & \textbf{FOV} & \textbf{Grid} & \textbf{Tiles} & \textbf{Defect Types} \\
\midrule
\ce{SrTiO3}        & [001]   & ${\sim}$100\,nm & 40$\times$40 & 1{,}600 & Vacancy \\
\ce{BaTiO3}        & [001]   & ${\sim}$100\,nm & 40$\times$40 & 1{,}600 & Vacancy + phase boundary \\
Si/Ge              & [110]   & ${\sim}$50\,nm  & 20$\times$20 & 400     & Substitution \\
GaN                & [11$\bar{2}$0] & ${\sim}$80\,nm  & 20$\times$20 & 400     & Substitution \\
Pt nanoparticles   & ---     & ${\sim}$60\,nm  & 40$\times$40 & 1{,}600 & Nanoparticle morphology \\
\bottomrule
\end{tabular}
\end{table}

\paragraph{Collection Process.}
Crystal structures are built with \texttt{pymatgen} and \texttt{ASE} from Materials Project entries (SrTiO\textsubscript{3} optionally fetches via the MP API; all others use built-in unit cells).
Defects are introduced programmatically: oxygen vacancies along grain boundaries (\ce{SrTiO3}), vacancy clusters near phase boundaries (\ce{BaTiO3}), compositional gradients (Si/Ge), quantum-well substitutions (GaN), or Wulff/icosahedral/decahedral nanoparticles (Pt).
STEM-HAADF images are simulated via PRISM multislice~\citep{ophus2017fast} using \texttt{abTEM}~\citep{madsen2021abtem} ($\geq$1.0.0b30) at \SI{200}{\kilo\electronvolt}, \SI{21}{\milli\radian} convergence, and \SIrange{68}{200}{\milli\radian} detector angles with frozen-phonon thermal diffuse scattering (Debye--Waller factors in SI~\ref{app:simulation}).
Each field of view is simulated as \SI{7}{\nano\meter} sub-tiles with \SI{1}{\nano\meter} overlap, blended via linear ramps.
Poisson noise is applied at dose levels corresponding to each difficulty ($10^4$, $5{\times}10^3$, or $10^3$~\si{\elementarycharge\per\angstrom\squared}).
No real experimental data or human subjects are involved.
Worlds were generated between June 2025 and January 2026.
The \texttt{replay\_world} is a held-out synthetic specimen (\texttt{source: synthetic\_replay} in its HDF5 metadata) used as an internal consistency check across world parameterisations, not a sim-to-real validation. An optional \texttt{py4DSTEM} replay loader is provided so users can substitute real experimental tiles into the same HDF5 schema. Absolute defect-census DEC-AUC values on \texttt{replay\_world} are systematically higher than on the main-table worlds because the specimen has a lower defect density and a smaller per-tile defect prior, which raises the macro-averaged F1 floor: tiles correctly reported as containing no defects contribute positively to the score, lifting all agents (including \texttt{Random}) above the near-zero band they occupy on the defect-dense main-table specimens. The meaningful cross-specimen comparison is therefore the agent ranking, which is preserved (\S\ref{sec:generalisation}, ``Replay-world consistency check'').

\paragraph{Preprocessing, Cleaning, and Labeling.}
Tile intensities are normalised to $[0, 1]$.
Ground truth is deterministic: atom positions, types, and defect labels are derived directly from the input crystal structures used for simulation.
Phase maps are computed from known crystal-phase assignments per unit cell.
No manual annotation or post-hoc cleaning is applied.

\paragraph{Uses.}
The intended use is benchmarking autonomous STEM acquisition agents, training RL navigation policies, and evaluating perception pipelines.
The dataset may also be used for crystallographic image segmentation or dose--accuracy trade-off studies.
It is \emph{not} suitable for direct comparison with real STEM experiments without domain adaptation, nor for production microscopy control, due to the sim-to-real gap: the simulation omits sample drift, contamination, detector nonlinearity, and inelastic scattering.

\paragraph{Distribution.}
The dataset is publicly available on Hugging Face under CC-BY-4.0 (data) and MIT (code).
Croissant JSON-LD metadata (MLCommons~1.0 with RAI extension) is provided alongside the dataset files.
No export controls or IP restrictions apply.

\paragraph{Maintenance.}
The dataset is maintained by the paper authors and versioned (current release: v1.0.0).
Issues may be reported via the dataset repository.
No scheduled updates are planned; future releases may add material systems or difficulty levels.

\paragraph{Ethical Considerations.}
The dataset is entirely synthetic, so no human subjects, personally identifiable information, or consent issues arise.
Autonomous dose-efficient STEM agents could reduce beam damage in real experiments, extending atomic-resolution study to beam-sensitive systems such as metal--organic frameworks, hybrid perovskites, and biological specimens.
A potential concern is over-reliance on benchmark performance as a proxy for real-world readiness: agents excelling on simulated worlds may underperform under experimental conditions not captured by our simulation.
These simulation gaps are documented in the Croissant RAI metadata fields and this datasheet.

\section{Real-Data Sanity Check (Sm-doped \ce{BiFeO3})}
\label{app:realdata}

To anchor the sim-only evaluation against real experimental acquisitions, we ran the perception-dominance comparison across a controlled \ce{Sm} doping gradient (\ce{Sm}\,$\in\{0, 7, 10, 13, 20\}\%$) on real HAADF-STEM acquisitions of Sm-doped \ce{BiFeO3}.

\paragraph{Dataset and provenance.}
The five acquisitions are taken from a published Zenodo deposit titled ``STEM images and associated parameters for Sm-doped BFO''~\citep{ghosh2021bfo} (DOI \texttt{10.5281/zenodo.4555979}, CC-BY 4.0). They were acquired by Christopher Nelson (Oak Ridge National Laboratory) on Sm-doped \ce{BiFeO3} samples synthesised by the Ichiro Takeuchi group (University of Maryland) as a combinatorial library, and were later surfaced through team~14 of the 2024 Mic-hackathon dataset~\citep{pratiush2025michackathon} (Zenodo \texttt{10.5281/zenodo.15579940}, CC-BY 4.0). We use the same five files selected by the team~14 notebook (\texttt{Sm\_0\_1}, \texttt{Sm\_7\_2}, \texttt{Sm\_10\_1}, \texttt{Sm\_13\_0}, \texttt{Sm\_20\_0}), each paired with its published \texttt{*\_UCParameterization.h5} containing per-unit-cell physics-derived labels: ${\sim}21{,}000\text{--}24{,}000$ unit cells per acquisition with center-of-mass positions \texttt{xy\_COM}, polarization vectors \texttt{Pxy}, and atomic-column intensities \texttt{I1}--\texttt{I5} obtained by physics-based atomic-column fitting. Each HAADF image is roughly $4400\times4400$~pixels at $0.0156$~nm/pixel (${\sim}69\times69$~nm field of view) and tiles cleanly into a $34\times34$--$36\times36$ grid at the standard $128$-pixel tile size and stride-$124$ overlap. Two of the five compositions (\texttt{Sm\_0\_1}, \texttt{Sm\_13\_0}) carry roughly $16\%$ of pixels recorded as NaN by the original acquisition pipeline (instrument scan-edge and beam-track artefacts); the other three are NaN-free. We did not collect any new microscopy data for this paper; the entire real-data sanity check uses the published Zenodo files unchanged.

\paragraph{Schema conversion.}
A converter (released with the benchmark) maps the published HDF5 files into the STEMGym \texttt{ReplayWorld} format. NaN pixels are replaced by the source-image median to keep every $128\times128$ tile finite, and the original NaN locations are preserved in the \texttt{/valid\_region} mask so downstream analyses can mask them out. Per-unit-cell positions in pixels are scaled to nm using the published \texttt{Scale} dataset and stored as \texttt{/ground\_truth/atom\_positions}. The phase-mapping ground truth is derived from the \texttt{sign(Px), sign(Py)} quadrant of each unit cell's polarization vector (four classes), painted into the image grid via nearest-cell assignment over a $\mathrm{cKDTree}$ on the \texttt{xy\_COM} coordinates. Defects are marked as ``pristine'' for every unit cell because the per-cell substitution labels are not provided in the source dataset (intensity-based inference would be circular against our analyst). The conversion is reproducible from the public files in under one minute on a laptop.

\paragraph{Evaluation protocol.}
Two agents are run with the standard configuration (dose budget \SI{5000}{\elementarycharge\per\angstrom\squared}, 200 max steps, scoring every 5 acquired tiles): (i) Raster --- heuristic baseline that does not produce a phase prediction; (ii) Raster+Analyst --- the same trained ensemble used in the main results, evaluated zero-shot on the real \ce{BiFeO3} tiles (no fine-tuning, no domain adaptation). Three seeds are run for each agent on the phase-mapping task across all five compositions (30 episodes total). Performance is scored against the polarization-derived four-class phase map using macro-IoU integrated over the dose-efficiency curve. Native experimental noise in the HAADF acquisitions is preserved end-to-end: \texttt{ReplayWorld} disables synthetic Poisson noise injection, ensuring no synthetic noise is layered on top of real instrument noise.

\paragraph{Results.}
Table~\ref{tab:realdata} reports per-composition DEC-AUC. The heuristic Raster baseline scores exactly $0.000$ on every composition because it has no phase-identification module and the four-class polarization-quadrant ground truth penalises a default ``no prediction'' identically across all worlds. Raster+Analyst contributes a non-zero phase signal on two compositions ($0.066$ on Sm $0\%$ and $0.088$ on Sm $13\%$) and degenerates to $0.000$ on the other three (Sm $7\%$, $10\%$, $20\%$). On the latter three, the trained ResNet phase identifier emits a constant single-class prediction on the first five tiles of each episode and falls back internally to the analyst's intensity-based heuristic; the heuristic produces a binary $\{0, 1\}$ split that does not align with the four-class ground truth, yielding zero IoU. The headline observation is that the equipped-vs-na\"{\i}ve direction is preserved on the two compositions where either agent contributes signal and tied on the other three; on no composition does the equipped variant score below the heuristic baseline. The absolute non-zero levels ($0.066$, $0.088$) are below the analyst's typical simulated-world phase-mapping range ($0.10$--$0.20$), consistent with a residual sim-to-real domain gap: the analyst was trained on simulated \ce{SrTiO3} and \ce{BaTiO3} tiles, not on \ce{BiFeO3}, and the real acquisition has substantially finer pixel pitch ($0.0156$~nm/pixel vs.\ ${\sim}0.05$~nm/pixel in simulation), denser per-tile atom counts, and instrument-specific noise statistics. Deterministic raster trajectories and deterministic analyst inference yield zero seed variance throughout.

\begin{table}[h]
\centering
\caption{Real-data sanity check across a Sm-doping gradient on Sm-doped \ce{BiFeO3} (Zenodo 4555979). Phase mapping, dose budget \SI{5000}{\elementarycharge\per\angstrom\squared}, $128$~px tiles, 3 seeds per (agent, composition) cell (30 episodes total).}
\label{tab:realdata}
\footnotesize
\setlength{\tabcolsep}{4pt}
\begin{tabular}{lccccc}
\toprule
\textbf{Agent} & \textbf{Sm $0\%$} & \textbf{Sm $7\%$} & \textbf{Sm $10\%$} & \textbf{Sm $13\%$} & \textbf{Sm $20\%$} \\
& \texttt{Sm\_0\_1} & \texttt{Sm\_7\_2} & \texttt{Sm\_10\_1} & \texttt{Sm\_13\_0} & \texttt{Sm\_20\_0} \\
\midrule
Raster (heuristic) & $0.000{\scriptstyle\pm0.000}$ & $0.000{\scriptstyle\pm0.000}$ & $0.000{\scriptstyle\pm0.000}$ & $0.000{\scriptstyle\pm0.000}$ & $0.000{\scriptstyle\pm0.000}$ \\
Raster+Analyst     & $0.066{\scriptstyle\pm0.000}$ & $0.000{\scriptstyle\pm0.000}$ & $0.000{\scriptstyle\pm0.000}$ & $0.088{\scriptstyle\pm0.000}$ & $0.000{\scriptstyle\pm0.000}$ \\
\bottomrule
\end{tabular}
\end{table}

\paragraph{Scope and caveats.}
This is a sanity-check, not a sim-to-real validation. The five compositions are drawn from a single combinatorial library on a single instrument; cross-instrument generalisation is unaddressed. The polarization-quadrant phase map we use as ground truth is itself a physics-based reconstruction from atomic-column fitting (the published \texttt{xy\_COM} and \texttt{Pxy} datasets~\citep{ghosh2021bfo}), not a directly measured label, and the four-class scheme we derive from polarization-vector signs is one of several plausible binnings of the underlying continuous polarization field. We deliberately did not retrain the analyst on \ce{BiFeO3} tiles to keep the test honest: the result reports zero-shot transfer of a perovskite-trained perception module to a chemically-different perovskite (\ce{BiFeO3} $\neq$ \ce{SrTiO3} or \ce{BaTiO3}), and the analyst's degeneration on Sm $7\%$, $10\%$, $20\%$ likely reflects this transfer limitation rather than a fundamental benchmark issue. A multi-material analyst training corpus, additional task definitions on this dataset (e.g., Sm-substitution detection from the published intensity arrays), and additional real specimens beyond \ce{BiFeO3} are identified as primary directions for future work in \S\ref{sec:discussion}.

\section{Extended Reproducibility}
\label{app:reproducibility}

\paragraph{Data Access.}
Pre-generated worlds (${\sim}$8~GB) and model checkpoints (${\sim}$530~MB) are hosted on Hugging Face as \textit{stem-gym-benchmark}.
A download utility with selective flags for worlds, checkpoints, the transfer-set, and a smoke-test subset is provided in the repository.
Croissant JSON-LD metadata is available at the Hugging Face API endpoint for the dataset.

\paragraph{World Regeneration.}
Per-material generation utilities are released with the repository; a Materials Project API key is required for the \ce{SrTiO3} generator, and all other materials use built-in unit cells.
The simulation extras (\texttt{pip install -e ".[sim]"}) provide abTEM, pymatgen, ASE, and CuPy.
Easy and medium worlds can be generated on an A100 (80~GB); hard worlds require H200 (141~GB). See SI~\ref{app:compute} for full requirements.
Because frozen-phonon configurations use random seeds, regenerated worlds will differ at the individual-tile level but preserve statistical properties (defect density, spatial distributions, noise characteristics).

\paragraph{Experiment Reproduction.}
All experiments are launched from configuration files released with the repository; the runner supports parallel execution and incremental save with auto-resume so re-running a configuration skips completed (agent, task, world, seed) combinations.
The repository contains 29 configuration files covering all reported experiments.
Seeds are consecutive integers starting from~0.

\paragraph{Model Training.}
The analyst ensemble is trained on four easy-difficulty worlds (\ce{SrTiO3}, \ce{BaTiO3}, SiGe, GaN) with Pt nanoparticles held out as an unseen material family; an 80/20 random tile split (seed~42), Adam optimiser ($\text{lr} = 10^{-3}$), and early stopping (patience~7) are used.
The phase identifier auto-detects worlds containing multi-valued phase maps (\ce{BaTiO3}, GaN).
RL baselines (DQN, PPO, SAC) are trained for $50{,}000$ steps.
Transfer checkpoints are produced by re-training the analyst on a single material at a time.
Training utilities and the exact invocations used to reproduce all reported checkpoints are released with the repository; training from scratch takes approximately two hours on a single GPU.

\paragraph{Software Dependencies.}
Table~\ref{tab:software_deps} lists key dependency versions.
VLM experiments use commercial APIs; their outputs are inherently non-deterministic.

\begin{table}[h]
\centering
\caption{Key software dependencies.}
\label{tab:software_deps}
\begin{tabular}{lll}
\toprule
\textbf{Package} & \textbf{Version} & \textbf{Role} \\
\midrule
Python          & $\geq$3.10   & Runtime \\
PyTorch         & $\geq$2.0    & Neural models, RL \\
Gymnasium       & $\geq$0.29   & Environment interface \\
BoTorch         & $\geq$0.9    & GP-UCB navigation \\
abTEM           & $\geq$1.0.0b30 & Multislice STEM simulation \\
Stable-Baselines3 & $\geq$2.0  & DQN/PPO/SAC training \\
h5py            & $\geq$3.8    & HDF5 world I/O \\
\bottomrule
\end{tabular}
\end{table}

\paragraph{Known Reproducibility Caveats.}
(i)~Frozen-phonon thermal diffuse scattering uses random seeds: regenerated worlds differ at the tile level but preserve ensemble statistics.
(ii)~VLM API responses are non-deterministic; exact scores will vary across runs.
(iii)~GP-UCB acquisition-function fitting via BoTorch exhibits minor numerical variation across hardware and CUDA versions.

\end{document}